%% file: 0_main.tex
\documentclass{article}
\usepackage{amssymb}
\usepackage{booktabs}

\usepackage[preprint]{corl_2025} 

\usepackage{graphicx}
\usepackage{wrapfig}
\usepackage[table,xcdraw,dvipsnames]{xcolor}
\usepackage{caption}
\usepackage{amsmath}

\usepackage[most]{tcolorbox}
\usepackage{caption}
\usepackage{lipsum} 
\usepackage{mdframed}
\usepackage{algorithm}
\usepackage{algpseudocode}
\usepackage{pifont}

\usepackage{booktabs}
\usepackage{array}
\usepackage{pifont}   
\usepackage{amssymb}  
\usepackage{graphicx} 

\newcommand{\ph}[1]{{\textbf{#1}:}} 

\title{Enter the Mind Palace: Reasoning and Planning for Long-term Active Embodied Question Answering}

\author{
  Authors\\
  Field AI Research Institute (FAIRI), SISL\\ 
  United States\\
  \texttt{author@fieldai.com} \\
}

\author{
\textbf{M. Fadhil Ginting}$^{1,2}$, 
\textbf{Dong-Ki Kim}$^{1}$, 
\textbf{Xiangyun Meng}$^{1}$, 
\textbf{Andrzej Reinke}$^{1}$, 
\textbf{Bandi Jai Krishna}$^{1}$, \\
\textbf{Navid Kayhani}$^{1}$,
\textbf{Oriana Peltzer}$^{1}$, 
\textbf{David D. Fan}$^{1}$, 
\textbf{Amirreza Shaban}$^{1}$, 
\textbf{Sung-Kyun Kim}$^{1}$, \\
\textbf{Mykel J. Kochenderfer}$^{2}$,
\textbf{Ali Agha}$^{1}$, 
\textbf{Shayegan Omidshafiei}$^{1}$ \\
\\
$^{1}$Field AI, 
$^{2}$Stanford University
\\
{\color{blue} \href{https://mind-palace-laeqa.github.io/}{mind-palace-laeqa.github.io}}
}

\begin{document}
\maketitle

\begin{abstract}
As robots become increasingly capable of operating over extended periods—spanning days, weeks, and even months—they are expected to accumulate knowledge of their environments and leverage this experience to assist humans more effectively. This paper studies the problem of Long-term Active Embodied Question Answering (LA-EQA), a new task in which a robot must both recall past experiences and actively explore its environment to answer complex, temporally-grounded questions. Unlike traditional EQA settings, which typically focus either on understanding the present environment alone or on recalling a single past observation, LA-EQA challenges an agent to reason over past, present, and possible future states, deciding when to explore, when to consult its memory, and when to stop gathering observations and provide a final answer. Standard EQA approaches based on large models struggle in this setting due to limited context windows, absence of persistent memory, and an inability to combine memory recall with active exploration. To address this, we propose a structured memory system for robots, inspired by the mind palace method from cognitive science. Our method encodes episodic experiences as scene-graph-based world instances, forming a reasoning and planning algorithm that enables targeted memory retrieval and guided navigation. To balance the exploration-recall trade-off, we introduce value-of-information-based stopping criteria that determine when the agent has gathered sufficient information. We evaluate our method on real-world experiments and introduce a new benchmark that spans popular simulation environments and actual industrial sites. Our approach significantly outperforms state-of-the-art baselines, yielding substantial gains in both answer accuracy and exploration efficiency.
\end{abstract}


\input{1_intro}
\input{2_rel_works}
\input{3_problem_formulation}
\input{4_mind_palace_exploration}
\input{5_benchmark}
\input{6_experiments}
\input{7_conclusions}
\input{8_limitation}


\clearpage


\bibliography{ref}  

\newpage
\input{9_appendix}

\end{document}

%% file: 1_intro.tex
\section{Introduction}

\begin{figure*}[t]
    \centering
    \includegraphics[width=0.99\linewidth]{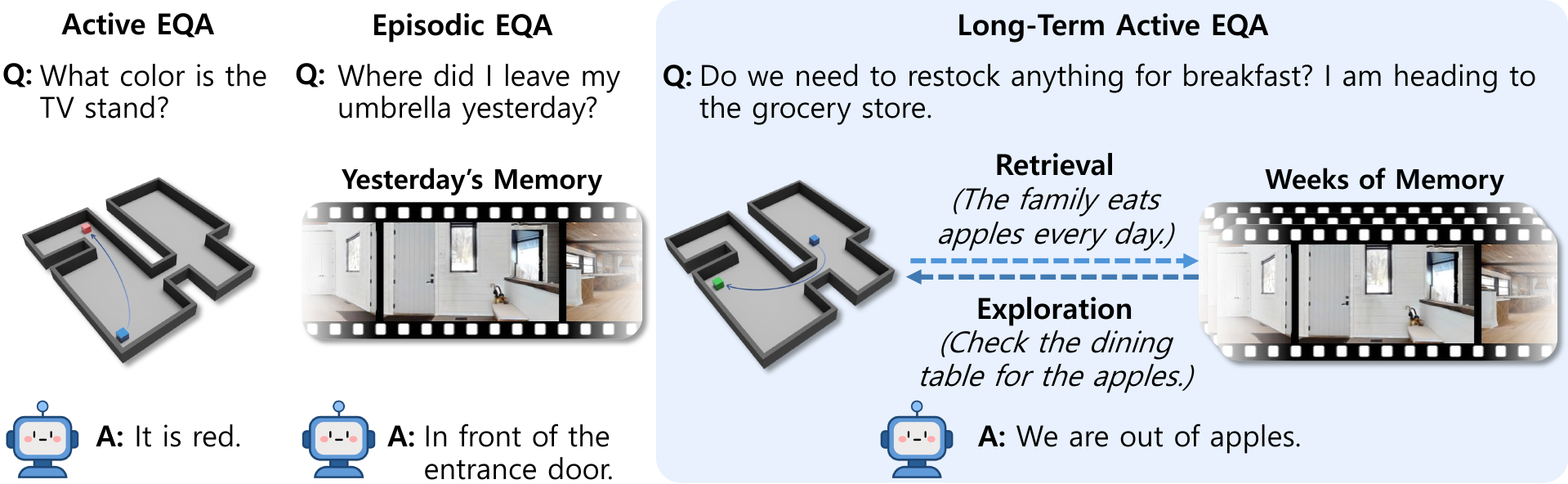}
    \caption{
    Different EQA problem setups. We study a new problem of Long-term  
Active EQA that combines active exploration with long-term memory understanding over multiple episodes.    } 
    \label{fig:figure1}
    \vspace{-0.7cm}
\end{figure*}

Humans naturally develop long-term situational awareness through repeated interactions with their environment, remembering routines, recognizing object placements, and anticipating future needs. 
For example, when making a shopping list for breakfast, one can recall household preferences and check available supplies to identify what needs to be bought.
This type of memory retrieval and long-term temporal grounding is key to intelligent embodied behavior.
Among tasks related to this, Embodied Question Answering (EQA) is particularly compelling, as it probes a robot's semantic understanding of its environment~\cite{das2018embodied}.
EQA approaches are typically framed either in active settings—where robots explore the environment from scratch to gather information~\cite{ren2024explore}—or episodic settings—where robots answer questions using a single recorded trajectory~\cite{majumdar2024openeqa}.
While Vision-Language Models (VLMs) have improved performance~\cite{liu2024improved,hurst2024gpt,grattafiori2024llama}, current approaches are limited to using only the robot's present observations or a single episodic memory, and do not generalize to using multiple past experiences or long-term knowledge. 
To address this gap, we introduce Long-term Active Embodied Question Answering (LA-EQA), where robots must both recall past experiences and actively explore their surroundings to answer complex questions~(see \autoref{fig:figure1}).
To our knowledge, this problem is largely unexplored, and no benchmark currently exists to evaluate it.

Performing LA-EQA with VLMs and LLMs using existing EQA approaches is challenging for two reasons.
First, representing the robot’s past observations accumulated over many deployments across days or months is difficult: a single run can generate thousands of images from diverse viewpoints, yet most questions only require a few relevant frames. Ingesting all this data directly is inefficient and often infeasible due to limited context windows. 
Second, retrieving relevant information from long-term memory and exploring relevant places in the environment creates a vast combined search space of past and new observations, where an uninformed search is computationally costly. These challenges raise a need for a new paradigm for long-term reasoning for embodied agents.

To address these challenges, we propose an approach for effective long-term memory representation and retrieval for embodied agents.
Inspired by the mind palace technique~\cite{legge2012buildingmindpalace}—where humans can effectively recall memories by associating them with spatial landmarks—we structure a robot's long-term observations into a series of spatial world instances.
Each instance is represented by a hierarchical scene graph that spatially groups semantic observations.
Spatiotemporal structure is captured by linking multiple episodic world instances over time, enabling reasoning and exploration using retrieval of relevant experiences based on spatial proximity and temporal context.
Our method, titled Mind Palace Exploration, has three components:
1) Generation, converting long-term memory into multiple scene-graph world instances;
2) Reasoning and Planning, where the robot interleaves EQA reasoning to identify target objects and assess if sufficient information has been gathered;
and 3) Stopping Criteria, using Value-of-Information to balance memory recall and active exploration.


We introduce the first benchmark on LA-EQA and evaluate our approach against state-of-the-art baselines in EQA. In particular, the benchmark consists of diverse large-scale, high-fidelity simulation environments and real-world office and industrial sites across multiple days and months. 
Our approach outperforms baselines by 12–28\% in answer correctness, achieves 16\% higher exploration efficiency, and maintains a 12\% correctness gain over the strongest baseline while using 77\% fewer retrieved images, demonstrating both the effectiveness and efficiency of our approach. 
We further demonstrate the scalability and generalizability of our method in long-term settings across diverse environment types, including reasoning over memory from deployments spanning 2.4 km of robot trajectories collected over 6 months.
We show the benefit of early memory retrieval stopping criteria in further reducing the number of past observation images while maintaining comparable performance. 
The real-world experiments demonstrate the feasibility of our approach in practical settings, where a legged robot deployed in a 1,000 $m^2$ office space uses past inspection memory to efficiently explore the environment and answer practical, day-to-day questions about the office.

%% file: 2_rel_works.tex
\section{Related Work}
\label{sec:related_works}

\textbf{Embodied Question Answering (EQA)}
has been studied extensively from earlier works that employed learning-based models~\cite{das2018embodied, gordon2018iqa, das2018neural, wijmans2019embodied, yu2019multi, thomason2019shifting} to more recent efforts leveraging foundation models~\cite{majumdar2024openeqa}.
Recent approaches generally fall into two settings: episodic-memory EQA, where the agent accesses a single episode of memory, such as in OpenEQA~\cite{majumdar2024openeqa} and ReMEmbR~\cite{anwar2024remembr}, and active EQA~\cite{jiang2025beyond,dorbala2024s,wu2024noisyeqa}, where the agent explores a novel environment to gather information for answering questions, such as in Explore-EQA~\cite{ren2024explore}, Efficient-EQA~\cite{cheng2024efficienteqa}, and Graph-EQA~\cite{saxena2024grapheqa}.
We propose a new and more general problem of Long-term Active EQA, in which the agent must integrate information across multiple prior episodes and active exploration to answer the question.


\textbf{Semantic scene representation} is a critical component for embodied reasoning and planning.
Various methods have been proposed to encode the semantics and contextual structure of the world, including dense 3D representations~\cite{peng2023openscene,shafiullah2022clip}, voxel maps~\cite{liu2024dynamem}, and scene graphs \cite{armeni20193d,rosinol2021kimera,werby2024hierarchical}. 
In our work, we opt for a scene graph approach~\cite{rana2023sayplan}, which has demonstrated effectiveness in EQA tasks~\cite{xie2024embodiedrag,saxena2024grapheqa, yang2024snapmem}, and can be integrated with scalable memory retrieval and planning. 
We extend the scene graph from a single environment snapshot to a series of episodic scene graphs labeled by macro-temporal intervals (e.g., hours, days), enabling the agent to reason over multiple world instances that capture how the environment evolves across long-term deployments.

\textbf{Semantic-guided navigation} focuses on reasoning and planning methods for robot navigation directed by semantic cues, which has a rich body of literature~\cite{anderson2018evaluation, deitke2022retrospectives} involving tasks specified by images~\cite{zhu2017target,mezghan2022memory}, object categories~\cite{yokoyama2024vlfm,Ginting2024Seek}, and natural language~\cite{chiang2024mobility,fu2024autoguide,ginting2024saycomply, long2024instructnav}. 
Our work related to semantic-based planning to search objects~\cite{khanna2024goat} and gather information for EQA taks~\cite{cheng2024efficienteqa}. 
The problems are typically framed as either online planning, which builds representations incrementally during execution~\cite{kim2021plgrim,zhao2023large, chen2023not}, or offline planning, which relies on pre-constructed maps of the environment~\cite{chen2023openvocab,gu2024conceptgraphs}. 
We address the challenge of leveraging multiple historical maps for online planning in long-term settings where the environment evolves over time. We propose a unified approach integrating offline memory retrieval with online exploration for LA-EQA.

%% file: 3_problem_formulation.tex
\section{Problem Formulation of Long-term Active EQA}
\label{sec:problem_formulation}
LA-EQA is a setting where an agent answers questions about the environment by actively exploring it and retrieving relevant information from long-term memory. 
The LA-EQA task is defined as tuple $(Q, M, E, x_0, A^*)$, where $Q$ is the question, $M=[m_1,\cdots,m_N]$ is a list of episodic memories, $E$ is the current environment, $x_0$ is the initial robot pose, and $A^*$ is the ground truth answer. 
The environment is dynamic: its visual appearance and object states can change over time. 
Each episodic memory $m_i=[m_{i,1},\cdots,m_{i,L}]$ contains $L$ tuples of past robot pose and image observations $m_{i,j}=(x_{i,j},o_{i,j})$ collected within a specific macro-temporal interval (e.g., hours). 
This formulation is a generalization of active EQA and episodic memory EQA with a single episode (see \autoref{appendix:problem-formulation}).



In LA-EQA, the agent follows policy $\pi(a_k \mid x_k, h_k, Q)$, mapping its state $x_k$ at time step $k$, working memory $h_k$ (history of action and observation since receiving $Q$), and the question to one of three possible actions: \textit{retrieve}, \textit{explore}, and \textit{answer}. 
The \textit{retrieve} action $a^R$ recalls a past memory $m_{i,j}$ into $h_k$. 
The \textit{explore} action $a^E$ moves the robot to viewpoint $w_i$ in $E$, storing the new observation $o_k$ in $h_k$; $w_i$ need not be near the robot and can be any obstacle-free space informed by prior experience.
The \textit{answer} action $a^T$ generates an answer $A$ in natural language based on $h$ and terminates the task. 

%% file: 4_mind_palace_exploration.tex
\section{Mind Palace Exploration for Solving Long-term Active EQA} 
\label{sec:methods}
\begin{figure*}[t]
    \centering
    \includegraphics[width=0.99\linewidth]{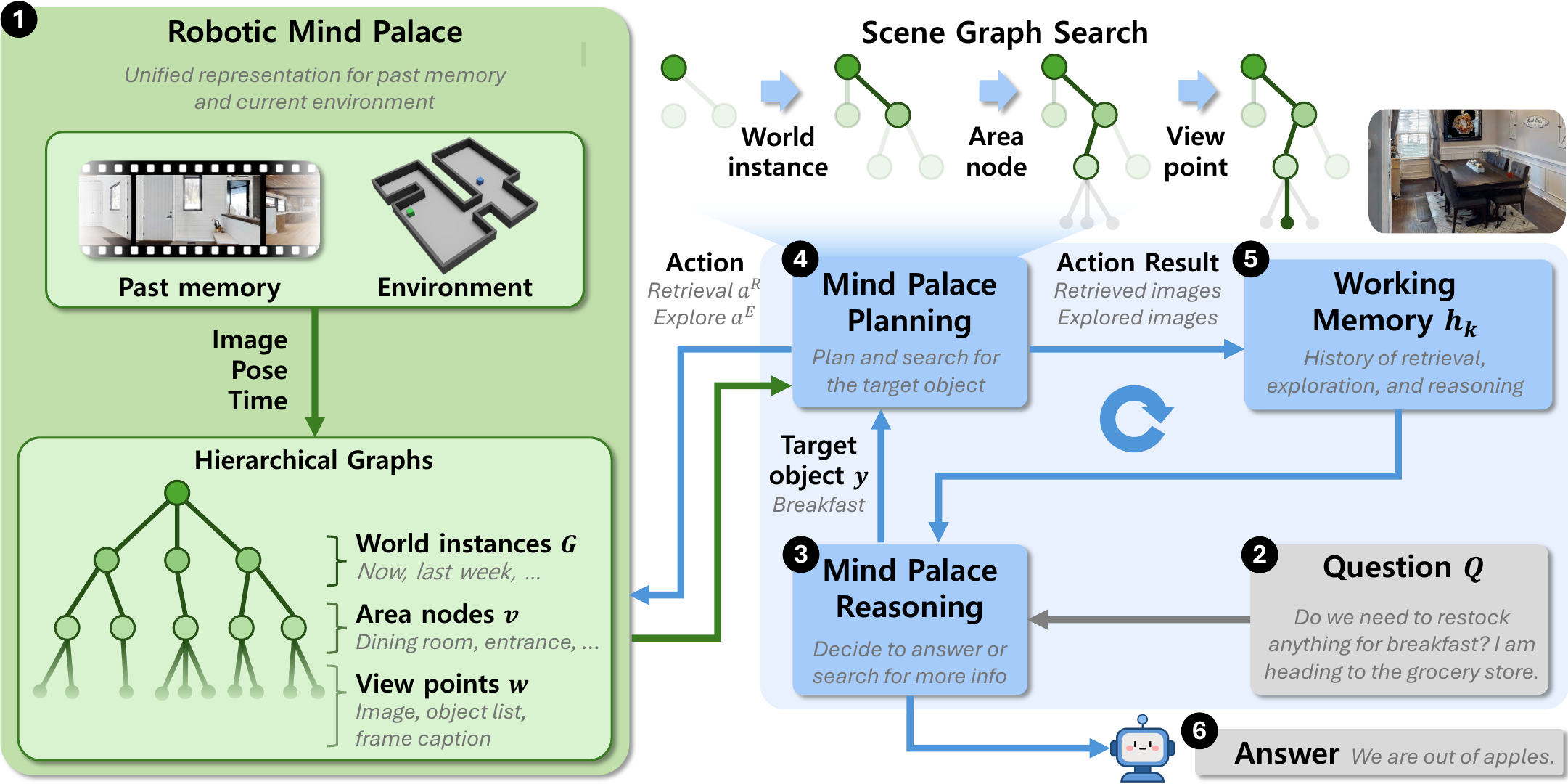}
    \caption{\textbf{Mind Palace Exploration} builds a Robotic Mind Palace that unifies past memories and environment representation \textit{(1)}. Given a question \textit{(2)}, the agent alternates between reasoning over the question to identify a target object \textit{(3)}, planning a search strategy through memory retrieval and exploration \textit{(4)}, and updating its working memory \textit{(5)}, until it is ready to answer the question \textit{(6)}.} 
    \label{fig:figure2_overview}
    \vspace{-0.5cm}
\end{figure*}

Humans use the mind palace technique~\cite{legge2012buildingmindpalace} to remember complex information by organizing it into a structured spatial memory, which enables efficient retrieval and traceable recall of relevant memories. We explore how this technique can be applied to long-term memory representation and reasoning in robots.
Our approach consists of three key ideas (see \autoref{fig:figure2_overview}). 
First, prior to the EQA task, we construct a long-term memory representation (referred to as the Robotic Mind Palace $\mathcal{M}$), which summarizes the robot’s history of observations into multiple world instances of scene graphs $\mathcal{M}=[G_0, G_1, \cdots, G_N]$. 
Then, during the LA-EQA scenario, the agent reasons over and explores these world instances in $\mathcal{M}$ to answer the question $Q$ using a policy $\pi$. 
Additionally, we introduce early stopping criteria using the notion of \textit{value of information} to avoid retrieving memory that is unlikely to improve the next exploration action $a^E$. We provide a detailed algorithm in \autoref{appendix:method-details}.


\subsection{Mind Palace Generation}

\textbf{Mind Palace is a series of episodic world instances. } 
The Mind Palace divides the long-term history of robot image observation and trajectories $M$ into episodes $m$ based on a macro-temporal term such as hours, times of day, and weeks.
The chunking of the episodes comes naturally in robotics as a mobile robot in continuous operations needs to pause any activities while recharging the battery. 
Each episode becomes a world instance in the Mind Palace and is indexed by its macro-temporal label in texts, allowing an LLM-based agent to select relevant episodes to recall. 

\textbf{An Episodic world instance is represented as a hierarchical scene graph. }
Given the sequence of robot observation and trajectory within an episode $m_i$, we build a world representation as a hierarchical scene graph $G_i=(\mathcal{V}_i,\mathcal{E}_i)$, where $\mathcal{V}_i$ denotes the set of nodes and $\mathcal{E}_i$ denotes the edges connecting the nodes~\cite{rosinol2021kimera}. 
First, we sample dense viewpoints $w$ from the past trajectory to form a set of viewpoint nodes. 
Each viewpoint node $w_i$ is associated with the robot pose $x$, images, a list of detected objects in the image~\cite{zhang2024tag,zhang2023recognize}, and frame captions. 
The list of objects and frame captions is used as an index for LLM-based agents for image retrieval selection. 
Then the viewpoint nodes $w$ are clustered into area nodes $v\in\mathcal{V}_i$ based on the spatial and contextual similarity~\cite{multi_hydra, xie2024embodiedrag, kayhani2023semantic}. 
Each area node $v$ is associated with the centroid of all the clustered viewpoints and the object list. 
The neighboring viewpoints $w$ and areas $v$ are connected with graph edges, and every $w$ is connected to a $v$, forming a hierarchical scene graph representation $G$ for each world instance.



The Robotic Mind Palace consists of a series of world instances representing the \emph{past} long-term memory $[G_1, \cdots,G_N]$ and the \emph{present} knowledge of the environment $G_0$. 
At the start of the LA-EQA task, we assume the robot has not explored the present environment yet, so world instance $G_0$ is only initialized with area nodes $v$ because the state of the environment and object placement may have changed since the last mapping in $G_1$. 
We update $G_0$ as the robot explores the environment. 


\subsection{Mind Palace Reasoning and Planning} 
We perform reasoning and planning over the robotic mind palace to solve the LA-EQA task. 
This involves three interleaving steps:
1) reasoning over the question to determine what object or spatial concept $y$ to search and when the agent can answer the question, 2) hierarchical planning over the Mind Palace to gather information, and 3) updating the information to the working memory $h$.

\ph{Reasoning over the LA-EQA}
The first step in the reasoning process is to determine whether the robot has sufficient information to answer question $Q$ using working memory $h_k$, which stores past actions, observations, and prior reasoning steps.
The agent queries a VLM with $h_k$ and $Q$. 
If the VLM responds it is possible to answer the question, the agent executes the \textit{answer} action and provides an answer $A$ with a VLM query. 
Otherwise, the agent queries and LLM to identify a target object or a spatial concept $y$, either a specific object explicitly stated in the question or an inferred cue (e.g., something to make a coffee), which becomes the next object goal for exploration. 

\ph{Planning over episodic world instances $G$ in the Mind Palace}
Mind Palace planning begins by selecting a sequence of world instances $G$ to locate $y$ efficiently. 
We query an LLM with a two-step reasoning process because we observe direct query often yields inefficient plans. 
The first step asks the LLM to reason whether answering the question requires object search across multiple world instances or if it only concerns a specific instance. 
Based on this reasoning, the LLM selects a subset of $G\in\mathcal{M}$ and plans a sequence of $G$. 
We guide the sequential planning with a heuristic that suggests prioritizing past world instances over the present instance $G_0$ as using prior knowledge of $y$ locations in the past can inform and improve object search efficiency in the present. 


\ph{Planning over areas $v$ in the scene graph}
Given a world instance $G_i$, we plan a sequence of areas $v$ to explore that maximizes the probability of finding $y$. 
This is framed as an object-goal navigation problem, and we adopt the planning formulation of object search over a scene graph~\cite{Ginting2024Seek}. 
We first query an LLM to output the probability of finding object $y$ on each area $v \in G_i$, 
then use a forward search planner to find the best sequence of areas $v$ to explore that minimizes the cost $J$ to find $y$~\cite{kochenderfer2022algorithms}. 
When exploring the present scene graph $G_0$, the cost is defined by the path length between the robot's current pose $x_k$ and the centroid of each area.
In contrast, when reasoning over past graphs, the agent can teleport to any area at a constant cost, regardless of the travel distance.

\ph{Exploring viewpoints $w$ and replanning}
Given an area $v_i$ to search, we query the LLM to select viewpoints $w$ based on the textual information in $G_i$. 
The object $y$ may appear in frame captions or the object list but often is not mentioned, and relevant viewpoints must be inferred given the textual information~\cite{shah2023navigation}. 
The robot then explores the viewpoints by recalling images from the Mind Palace or navigating to the viewpoints in the environment using a robot-specific motion planner and taking the images.
The retrieved or observed images are then stored in the working memory $h_k$. 
We repeat the planning over areas $v$ and viewpoints $w$ until the object $y$ is detected in images by a VLM or until we reach the exploration limits. 
If the object is detected, we search for $y$ in remaining world instances $G$ and move to the \textit{reasoning over the LA-EQA} step.



\subsection{Early Stopping of Memory Retrieval for Navigation}
This section examines how to reduce memory retrieval while maintaining exploration efficiency comparable to that of the unlimited memory retrieval case.
In particular, we develop stopping criteria that decide when to halt past memory retrieval and proceed with exploration. 
Given a sequence of world instances that includes the present instance $G_0$ (e.g., $[G_1, G_2,G_0]$), we use an LLM to form a prediction set of areas $v\in G_0$, where the object $y$ can be located with a probability above a threshold $P(y) \geq 1-q$. 
Studies have shown that the LLM prediction and threshold $1-q$ can be calibrated~\cite{ren2024explore, ren2023robots,tian2023just}.  
Using the prediction set, we define two possible conditions to immediately stop memory retrieval from past world instances $[G_1, G_2]$: 1) the prediction set contains only one area; 2) further memory retrieval will not improve the robot plan over the next sequence to explore $v_i$ in the prediction set. 
We evaluate the possible improvements on the sequence using the notion of Value of Information (VoI)~\cite{howard2007information}, which quantifies the expected utility gain from retrieving past memory, reducing the expected exploration cost $J$. We discuss more details in~\autoref{appendix:method-details}.

%% file: 5_benchmark.tex
\section{Long-term Active EQA Benchmark} 

\begin{figure*}[t]
    \centering
    \includegraphics[width=1.0\linewidth]{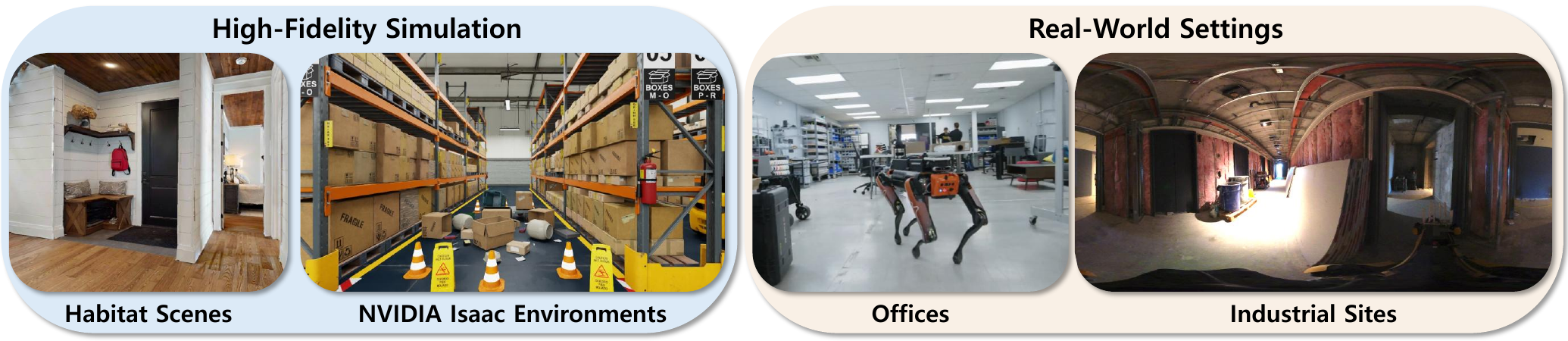}
    \caption{\textbf{LA-EQA Benchmark:} simulated and real-world scenes spanning multiple days / months.}
    \label{fig:figure3_benchmark}
    \vspace{-0.5cm}
\end{figure*}

\label{sec:dataset}
Existing EQA datasets~\cite{majumdar2024openeqa,ren2024explore,anwar2024remembr, jiang2025beyond} focus on scene understanding over short time spans (i.e., the same day), limiting their ability to capture long-term evolution of a scene (e.g.,  days and months). To address this, we curate the first LA-EQA dataset and benchmark, consisting of 3 simulated and 2 real-world scenes (see \autoref{fig:figure3_benchmark}). 
For each simulation scene, we generate 5--10 scene variations over multiple days, reflecting changes caused by common routines. 
For real-world scenes, we collected 11 trajectories (30--60 mins) in an industrial site and an office environment over a 6-month period. 

\ph{Question types} 
We categorize the questions based on their required temporal reasoning to capture different aspects of long-term scene understanding. 
\textbf{1) Past questions} pertain to a specific event observed in a single past trajectory. 
\textbf{2) Present questions} require only exploration of the current environment. 
\textbf{3) Multi-past questions} involve synthesizing information from multiple past trajectories (e.g., ``What do we usually eat for breakfast?'').
\textbf{4) Past-present questions} require reasoning over both historical memory and the current scene (e.g., ``Are we missing anything we usually have for breakfast?'').
\textbf{5) Past-present-future questions} involve predicting future outcomes based on both past and present observations (e.g., ``When do you think we will run out of apples for breakfast?'').

We curated 150 questions, which uniformly cover the question types. The questions were generated by seven people to ensure the diversity of the questions. The dataset consists of past trajectories and observations, simulation environments, ground truth answers, and exploration solutions. More detailed statistics and representative questions are presented in \autoref{appendix:benchmark-details}.




%% file: 6_experiments.tex
\section{Experiments and Discussion} 
\label{sec:results}
To evaluate our method, we answer: \textbf{Q1)} Does Mind Palace Exploration outperform other EQA methods across question types and lengths of past memory in long-term active EQA? \textbf{Q2)} Does early stopping of memory retrieval reduce the amount of memory retrieved without sacrificing performance? \textbf{Q3)} Can Mind Palace Exploration be practically deployed in real-world settings?

\ph{Methods}
We compare our approach against the following baselines: 
\textbf{1) Multi-Frame VLMs} process the question with images and robot poses through a VLM to output the answer. This method is the strongest approach in the OpenEQA benchmark. \textbf{2) Socratic LLMs w/ Frame and Scene Graph Captions} use image and scene-graph captions and robot poses to answer the question. \textbf{3) ReMEmbR~\cite{anwar2024remembr}} is a state-of-the-art method in episodic EQA by building a queryable vector database representation of the robot pose, observation time, and image caption embedding and retrieving relevant entries in the database using an LLM. We use the open-source code of the method. \textbf{4) Active EQA Agent w/ Frames as the Memory} has the same information as Multi-Frame VLMs, but it lets the agent explore the environment by providing a list of viewpoints that the robot can visit. This approach is similar to the state-of-the-art method of using long-context VLMs with topological graphs~\cite{chiang2024mobility} applied to the LA-EQA setting. \textbf{5) Active Socratic EQA Agent w/ Captions as the Memory} uses the same past memory information as  Socratic LLMs w/ Frame and Scene Graph Captions, but it lets the agent explore viewpoints and analyze explored images with VLMs. All approaches use the GPT-4o as the language and vision model~\cite{hurst2024gpt} and have the same maximum budget image retrieval, wherever applicable, and have the same exploration budgets on all active methods. The implementation details are provided in~\autoref{appendix:baseline-setup}. 

\ph{Metrics} We evaluate all the agents using three metrics: \textbf{1) Answer correctness} is compared to the human-annotated answer and judged by an LLM-based scoring~\cite{majumdar2024openeqa}. \textbf{2) Exploration efficiency} measures the path length of robot exploration compared to the oracle path length weighted by \textit{answer correctness}. \textbf{3) Memory retrieval efficiency} measures the number of past images retrieved to answer the question. The details of the metrics are provided in~\autoref{appendix:evaluation-metrics}.

\begin{figure*}[t]
\centering

\begin{minipage}[t]{0.5\textwidth}
\vspace{-80pt}  
\centering
\footnotesize
\resizebox{\linewidth}{!}{%
\begin{tabular}{@{}lccc@{}}
\toprule
\textbf{Methods} & \textbf{Answer} & \textbf{Expl. Eff.} & \textbf{Mem. (\#)} \\ 
\midrule
\rowcolor{cyan!10} Mind Palace (Ours) & \textbf{65.0\%} & \textbf{0.45} & 22.86 \\
\rowcolor{cyan!10} Mind Palace w/ stopping & 61.8\% & 0.42 & \textbf{15.73} \\
Multi-Frame VLMs~\cite{majumdar2024openeqa} & 52.9\% & - & 100 \\
Socratic LLMs~\cite{majumdar2024openeqa} & 44.3\% & - & 0 \\
ReMEmbR~\cite{anwar2024remembr} & 46.1\% & - & 0 \\
Active EQA w/ Frames & 43.7\% & 0.29 & 100 \\
Active Socratic EQA & 36.8\% & 0.19 & 0 \\
\bottomrule
\end{tabular}
}
\captionof{table}{LA-EQA results over answer correctness, exploration and retrieval efficiency.}
\label{tab:comparison_results}
\end{minipage}
\hfill
\begin{minipage}[t]{0.49\textwidth}
\centering
\includegraphics[width=0.9\linewidth]{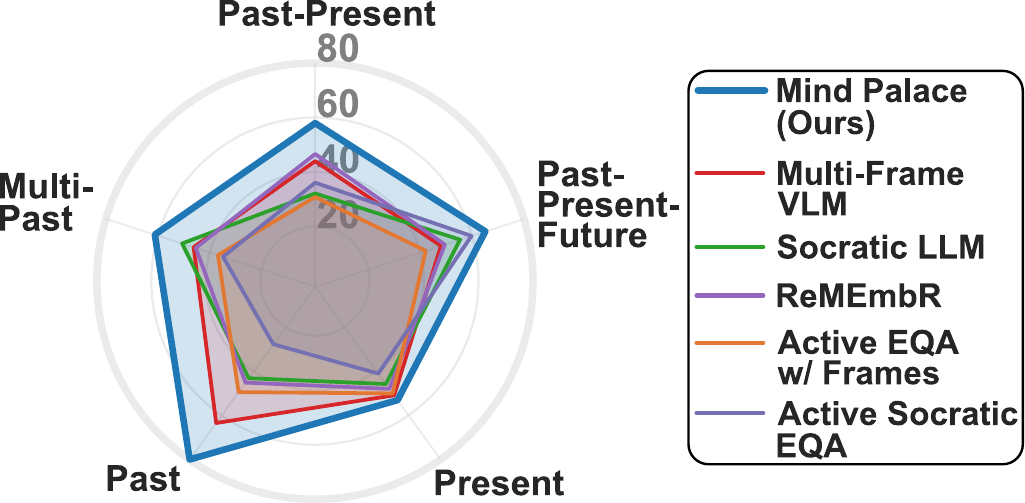}
\captionof{figure}{Performance over temporal reasoning question types.}
\label{fig:question_type_performance}
\end{minipage}
\vspace{-0.5cm}
\end{figure*}

\subsection{Q1) Mind Palace Exploration Outperforms other EQA Approaches}

\begin{wrapfigure}{r}{0.45\textwidth}
    \centering
    \includegraphics[width=\linewidth]{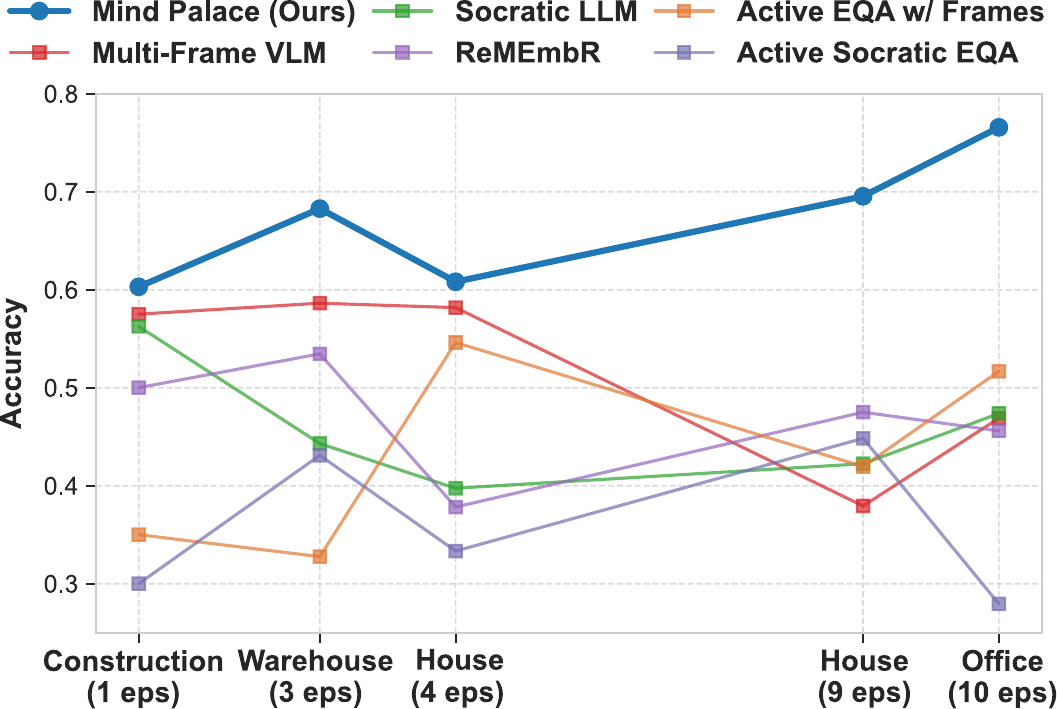}
    \caption{Performance of five different environments in the LA-EQA benchmark with a varying number of episodes.} 
    \label{fig:environment_type_performance}
    \vspace{-0.2cm}
\end{wrapfigure}

\textbf{Mind Palace Exploration outperforms baselines in all metrics.} As shown in Table \ref{tab:comparison_results}, our approach considerably outperforms all methods across the metrics, highlighting the gap in the current EQA approaches in the long-term EQA setting. 
More detailed results are provided in~\autoref{appendix:benchmark-more-results}.

\textbf{Efficient past image retrieval is the key to multi-episodic world understanding. }
Our approach significantly outperforms the others that require specific information from past memory, represented by \textit{past} and \textit{multi-past} question types in \autoref{fig:question_type_performance}. 
This is largely because images convey richer visual contexts than captions, enabling more accurate answers about object properties, states, and placements.
In the LA-EQA setting, multi-frame VLMs struggle as the maximum context length of the state-of-the-art VLMs is not comparable to the sheer amount of past observations in the memory. 
Our image retrieval approach is critical for efficient image analysis, as EQA questions typically need only several question-related images across multiple episodic memories. 
The results in Table \ref{tab:comparison_results} show that our approach only needs 77.14\% fewer images compared to VLM-based methods, with much higher answer correctness.

\textbf{Leveraging long-term memory improves active exploration efficiency. }
Our method achieves higher exploration efficiency than other active EQA agents (Table \ref{tab:comparison_results}), particularly on \textit{past-present} questions (\autoref{fig:question_type_performance}) in which past information can benefit present exploration. 
Our approach often recalls past memories to locate objects of interest so it can more accurately predict the probabilities of the object placements across areas $v$ in the present environment. 

\textbf{Mind Palace Exploration is a scalable approach for LA-EQA.} 
We evaluate scalability by plotting answer accuracy across different environments with varying numbers of past episodic memories (\autoref{fig:question_type_performance}). 
Our approach shows increasing performance gains over other methods with the number of past episodes in the memory. 
Given the same image retrieval limits, multi-frame VLM performance considerably drops as the images have less coverage across all the memories.  
ReMEmbR performs steadily, highlighting the value of retrieval-based approaches in long-term EQA problems.

\textbf{Our approach generalizes to diverse environments beyond the standard house setting.}
To test our approach further beyond standard EQA home environments benchmarks, we evaluated Mind Palace Exploration in larger real-world construction sites, a large office, and a simulated warehouse (\autoref{fig:environment_type_performance}), where it consistently outperforms others, highlighting its flexibility. 
Building a structured memory representation for efficient exploration and retrieval becomes more critical as the environment size increases across many episodes. See \autoref{appendix:full-examples} for full results. 




\subsection{Q2) Benefits of Early Memory Retrieval Stopping}


\textbf{Early memory retrieval stopping reduces the number of memories retrieved without sacrificing performance. }
As shown in Table \ref{tab:comparison_results}, early stopping reduces the amount of image retrieval from the past memory while maintaining comparable answer accuracy. 
The early stopping reduces the number of past world instances that the agent retrieves if there is no new observation that will change the agent's next exploration action. 
Examples in the experiment that we observe include when the robot predicts the possible areas where the object of interest is on the second floor, the robot will stop retrieving past world instances and move to the second floor. 
The stopping criteria are beneficial to even further improve the memory retrieval efficiency in Mind Palace Exploration.

\begin{figure*}[t]
    \centering
    \includegraphics[width=0.99\linewidth]{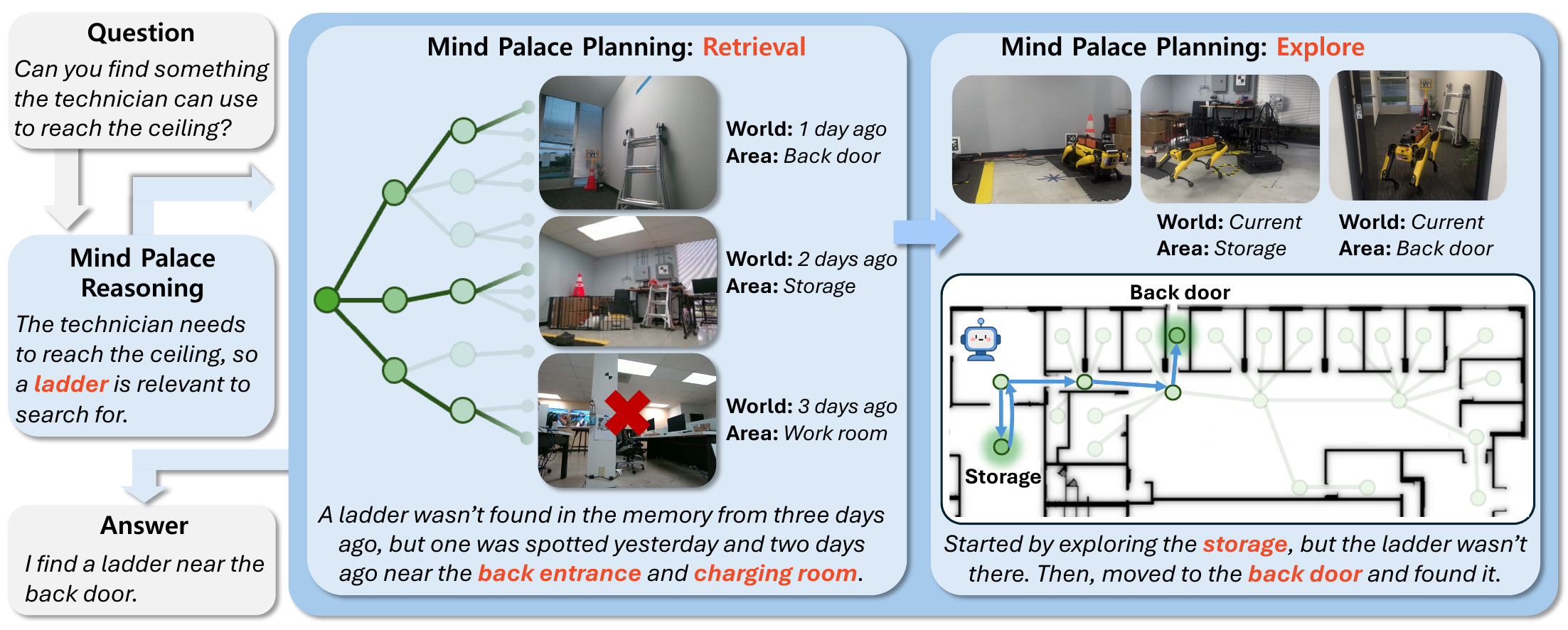}
    \caption{\textbf{Long-term Active EQA Hardware Experiments.} The robot retrieve relevant past information from the past memory and navigate around the office to answer the question.}
    \label{fig:figure6_hardwaew}
    \vspace{-0.5cm}
\end{figure*}

\subsection{Q3) Real-world Hardware Experiments}
We demonstrate the efficacy of Mind Palace Exploration in real-world LA-EQA use cases in an office space spanning over 1,000 $m^2$ with 27 different areas, using a legged robot as an office assistant. 
The robot accesses 10 past episodes of past runs, inspecting the office for the past four days and six monthly inspections from October 2024 to March 2025.
All the Mind Palace memory storage and planning, other than the GPT4-o query, is performed on the robot.
A user sends the question to the robot remotely through a computer, and the robot reports back the answer once it finishes the task. 
We select 7 questions from the LA-EQA benchmark that require active exploration~(\autoref{fig:figure6_hardwaew}).

\textbf{Mind Palace Exploration enables efficient exploration for practical real-world tasks. }
By consolidating knowledge of past object placements, the robot can efficiently locate relevant objects, saving an average of 3–10 room searches across the seven evaluated questions compared to a robot without memory access. 
The questions reflect realistic office scenarios (e.g., searching for tools, tracking missing packages, or identifying vacant desks unused for days) demonstrating the practical utility of LA-EQA. The robot can answer all the questions given that the information is available in its past memory and the current environment. See~\autoref{appendix:hardware-results} for hardware experiment details.

%% file: 7_conclusions.tex
\section{Conclusion}
\label{sec:conclusion}
We present the problem of LA-EQA, a new task that requires robots to combine long-term environment understanding with active exploration.
We propose Mind Palace Exploration to address LA-EQA by representing past long-term memory and the present environment with a robotic mind palace, enabling reasoning and planning over the Mind Palace.
We introduce the first benchmark for long-term active EQA, spanning days of simulation environments and months of real-world data, to foster future research in long-term reasoning.
Our approach significantly outperforms state-of-the-art EQA baselines, highlighting the need for a new paradigm for LA-EQA.

%% file: 8_limitation.tex

\section{Limitations}
\label{sec:limitations}
We discuss the limitations and challenges that we faced while evaluating our approach on the LA-EQA problem and building the Long-term EQA benchmark.


\textbf{Long-term understanding is limited to the coverage of past trajectories. }
When evaluating our approach in simulations, we design questions that a human could answer by reviewing past memories or by exploring the environment.
Our approach also assumes that questions can be answered using a combination of exploration and retrieval from past memories.
However, in real-world hardware experiments, we find a question that requires broader past coverage to be answered accurately.
The question asks the robot to find a vacant desk in the office that had not been used in recent days.
Our approach mistakenly selects an occupied desk because, in the robot's past data covering a one-hour walk every day, the desk is observed as vacant. 
This limitation highlights an exciting future research direction where robotic agents can recognize when they lack sufficient information and request additional information from humans or other agents.


\textbf{Answer correctness gap due to limitations in semantic understanding of images and spatial understanding over the memories and environment.} 
We observe that the cases where our approach does not achieve full scores on benchmark questions are primarily due to limitations of the VLMs' semantic and spatial understanding capabilities.
The semantic understanding issues include failures in detecting, counting, and understanding the functional properties of objects.
Stronger VLMs or multiple specialized vision models can be easily integrated into the Mind Palace Exploration framework, and we expect performance to improve as the VLMs' capabilities increase. 
Moreover, we observe several failed examples involving spatial understanding of images. 
For example, a VLM often assumes all objects detected while the robot is standing in one area are associated with the same area. This assumption does not hold when the images contain views of other areas. 
We find the most challenging questions for all approaches often require agents to collect multiple pieces of information and search different objects within the same environment and across multiple episodes. 
The problem highlights future research directions to further improve spatiotemporal reasoning capabilities for robots in a long-term setting, as measured by the performance in Long-term Active EQA benchmarks. 
We provide examples of the failure cases in \autoref{appendix:failure-modes}. 


\textbf{Long-term EQA benchmarks require manual design by humans, limiting scalability.}
Our benchmark is currently designed by a team of human experts, and the team carefully curates the questions. 
However, this approach limits the scalability of the benchmark generation.
We find several challenges in building the first long-term EQA benchmarks.
First, in the simulation environment design, a human must carefully curate long-term scenarios spanning multiple days and determine the changes in object placements over time.
We also need to place the objects manually, as we did not find reliable tools that can automate this process across different simulation worlds.
Moreover, the questions still need to be carefully designed by humans to ensure their quality and relevance to typical long-term EQA tasks in home and industrial use cases. 
This challenge becomes more pronounced when creating questions from extensive real-world data. 
We experimented with using LLMs for question generation, but the results were still far from effective.
Addressing these limitations in long-term EQA benchmark generation will require further research on automated scenario generation, scene modification, and practical long-term EQA question generation. 








%% file: 9_appendix.tex
\appendix
\section*{Project Webpage}
Our project website can be accessed at: \url{https://mind-palace-laeqa.github.io/} \\ (currently anonymized).

\section*{Appendix Contents}
\subsection*{Appendix A: Extended Problem Formulation of LA-EQA \dotfill \pageref{sec:appendixA}}
\subsection*{Appendix B: Mind Palace Exploration Details \dotfill \pageref{appendix:method-details}}
\subsection*{Appendix C: Long-term Active EQA Benchmark \dotfill \pageref{appendix:benchmark-details}}
\subsection*{Appendix D: EQA Methods Details \dotfill \pageref{appendix:baseline-setup}}
\subsection*{Appendix E: Evaluation Metrics \dotfill \pageref{appendix:evaluation-metrics}}
\subsection*{Appendix F: Example Results from Different Methods \dotfill \pageref{appendix:benchmark-more-results}}
\subsection*{Appendix G: Hardware Experiment Details \dotfill \pageref{appendix:hardware-results}}
\subsection*{Appendix H: Examples of Failure Cases \dotfill \pageref{appendix:failure-modes}}
\subsection*{Appendix I: Example of Full Results \dotfill \pageref{appendix:full-examples}}


\newpage
\input{Appendix/A_Problem_Formulation}\label{sec:appendixA}

\input{Appendix/B_Mind_Palace_Exploration_Details}

\input{Appendix/C_Long_Term_Active_EQA_Benchmarks}


\input{Appendix/D_Baseline_Details}
\input{Appendix/E_Evaluation_Metrics}
\input{Appendix/F_Benchmark_Results}

\input{Appendix/G_Hardware_Experiments}

\input{Appendix/H_Failure_Cases}

\input{Appendix/I_Full_Results}\label{sec:appendixI}


%% file: Appendix/A_Problem_Formulation.tex
\section{Extended Problem Formulation of LA-EQA}
\label{appendix:problem-formulation}

The Long-term Active EQA (LA-EQA) problem is a generalized formulation that integrates active EQA (A-EQA) and episodic memory EQA (EM-EQA)~\cite{majumdar2024openeqa}, and extends them to support reasoning over long-term, multi-episodic memories.

\textbf{Active EQA (A-EQA)}~\cite{ren2024explore,cheng2024efficienteqa} is defined by the tuple $(Q, E, x_0, A^*)$, 
where $Q$ is the question, $E$ is the current environment, $x_0$ is the initial robot pose, and $A^*$ is the ground truth answer. The robot has no knowledge of $E$ and must actively explore the environment until it determines it has enough information to produce an answer $A$.

\textbf{Episodic Memory EQA (EM-EQA)}~\cite{majumdar2024openeqa} provides the robot with a single episodic memory $m_i$, which contains a sequence of pre-recorded robot poses and image observations $m_{i,j} = (x_{i,j}, o_{i,j})$. The task is defined as the tuple $(Q, m_i, A^*)$, and the agent must answer the question based solely on the provided memory $m_i$.

\textbf{Long-term Active EQA (LA-EQA)} unifies the A-EQA and EM-EQA formulations by giving the robot access to both multiple episodic memories and the ability to explore the current environment. This setting better reflects real-world scenarios, where humans answer questions by combining present exploration with the recall of relevant past experiences. The past episodic memories may range from a single trajectory $m_i$ to a sequence of $N$ trajectories $M = [m_1, \cdots, m_N]$ accumulated over multiple days, weeks, or even months.

%% file: Appendix/B_Mind_Palace_Exploration_Details.tex
\section{Mind Palace Exploration Details}\label{appendix:method-details}

\subsection{Implementation Details of Mind Palace Generation }

\ph{Hierarchical scene graph generation} 
We sample the robot’s past trajectory to ensure maximal coverage while minimizing redundant images stored in the long-term representation.
In the simulation environment, viewpoints are sampled every three simulation time steps.
For real-world data, we sample every three seconds and remove viewpoints with identical poses when the robot is stationary.
We use GPT-4o to generate captions for the viewpoint images and to list the objects visible in each image.
The viewpoints are clustered into areas by querying VLMs using the viewpoint positions, captions, and images.
For real-world office and industrial data, where floor plan maps are available, we cluster viewpoints based on the areas defined in the floor plans.

\subsection{Implementation Details of Reasoning over the LA-EQA}
The following prompt is used to determine whether the robot has sufficient information to answer the question.

\begin{tcolorbox}[title=Prompt 1: Reasoning over LA-EQA to determine whether the robotic agent has sufficient information to answer the question,
colback=gray!1!white, colframe=blue!50!white, fonttitle=\bfseries, coltitle=black]
\texttt{You are an AI agent operating in an environment. Your task is to answer user questions by either exploring the environment or recalling relevant past information.} \\ \\
\texttt{The user has asked the following question:} \\ \\
\texttt{Question: \textcolor{green!50!black}{\{question\}}} \\ \\
\texttt{Here is a summary of the information collected so far from exploration or memory recall:} \\ \\
\texttt{\textcolor{green!50!black}{\{working\_memory\}}}
\end{tcolorbox}

\begin{tcolorbox}[title=Prompt 1 (continued),
colback=gray!1!white, colframe=blue!50!white, fonttitle=\bfseries, coltitle=black]
\texttt{Based on this information, do you think you have enough information to answer the question?} \\
\texttt{If not, explain why and identify which objects or areas still need to be explored.} \\ \\
\texttt{Respond with a brief explanation to support your answer.}
\end{tcolorbox}

If the agent believes it still needs to gather more information, the following prompt template is used to identify the target object or spatial concept. 

\begin{tcolorbox}[title=Prompt 2: Reasoning over LA-EQA to identify object of interest or spatial concept to search,
colback=gray!1!white, colframe=blue!50!white, fonttitle=\bfseries, coltitle=black]
\texttt{To answer the user's question, what specific objects or places in the house should the agent focus on?} \\
\texttt{Specify an object, contextual description, or entity that could be observed by a robot’s camera while walking around a house.} \\ \\
\texttt{Question: \textcolor{green!50!black}{\{user\_question\}}} \\ \\
\texttt{For context, here is a summary of the agent’s exploration and observations so far:} \\
\texttt{\textcolor{green!50!black}{\{working\_memory\}}} \\ \\
\texttt{It is important to explicitly include any object mentioned in the question within the answer. If the question does not mention a specific object but instead refers to an object by function or property (e.g., something that can be used for X or something with a certain color), respond with a more general object description rather than naming a specific item.} \\ \\
\texttt{For instance, if the question is “What can I use to make tea?”, respond with:} \\
\texttt{“something that can be used to make tea” — not with a specific answer like “kettle,” since the available objects are still unknown.} \\ \\
\texttt{Do not hypothesize the object's location at this stage. The goal is to identify objects or entities relevant to the question, not to guess where they are.} \\ \\
\texttt{Keep the object description under 10 words.}
\end{tcolorbox}

If the agent believes it has sufficient information, the following query is used to return the answer. 

\begin{tcolorbox}[title=Prompt 3: Reasoning over LA-EQA to answer the question,
colback=gray!1!white, colframe=blue!50!white, fonttitle=\bfseries, coltitle=black]
\texttt{The user has asked the following question:} \\
\texttt{Question: \textcolor{green!50!black}{\{question\}}} \\ \\
\texttt{We are now ready to answer the question based on the following working memory, which includes relevant observations and past exploration:} \\
\texttt{\textcolor{green!50!black}{\{working\_memory\}}}
\end{tcolorbox}

\begin{tcolorbox}[title=Prompt 3 (Continued),
colback=gray!1!white, colframe=blue!50!white, fonttitle=\bfseries, coltitle=black]
\texttt{Based on this information, please answer the user’s question directly. Provide a concise answer and reason to your answer.} 
\end{tcolorbox}

\subsection{Implementation Details Planning over episodic world instances}
\ph{Benefits of the two-step reasoning process} 
The following example highlights the need of a two-step reasoning process to query LLM to plan over an episodic world instances.
\begin{tcolorbox}[title=Example 1: Benefits of the two-step reasoning process in planning over episodic world instances,
colback=gray!1!white, colframe=green!50!white, fonttitle=\bfseries,coltitle=black]
\texttt{Question:} \texttt{Do you know where is my backpack? I forget where I put it.} \\ \\
\texttt{List of episodic world instances:} \texttt{['now', 'yesterday evening']} \\ \\
\texttt{Direct query:} \textcolor{orange!85!black}{\texttt{We need to search ['now'].}} \\ \\
\texttt{Two-step reasoning process:} \textcolor{green!50!black}{\texttt{Recalling the placement of the backpack from the memory will help to search the backpack in the present. We need to search ['yesterday evening', 'now'].}}
\end{tcolorbox}

We use the following prompt template for planning over episodic world instances $G$. 
\begin{tcolorbox}[title=Prompt 4: Planning over episodic world instances,
colback=gray!1!white, colframe=blue!50!white, fonttitle=\bfseries, coltitle=black]
\texttt{In this query, we aim to identify time instances—either through present exploration or past memory recall—} \\
\texttt{that are most relevant for finding the target object \textcolor{green!50!black}{\{y\}} in order to answer the user’s question.} \\ \\
\texttt{Currently, we are searching for the object: \textcolor{green!50!black}{\{y\_object\_to\_search\}}} \\ \\
\texttt{User question: \textcolor{green!50!black}{\{Question\}}} \\ \\
\texttt{Reason for searching this object: \textcolor{green!50!black}{\{y\_reasoning\_to\_search\_object\}}} \\ \\
\texttt{Available time instances: \textcolor{green!50!black}{\{list\_of\_world\_instances\_time\}}} \\ \\
\texttt{We want to reason about which of the following search strategies is most efficient:} \\
\texttt{1. PAST\_ONLY, Answerable using only past memory.} \\
\texttt{2. PRESENT\_ONLY, Only present exploration is relevant.} \\
\texttt{3. PAST\_THEN\_PRESENT — Use past memory first to guide efficient current search.} \\
\texttt{4. MULTI\_PAST\_AND\_PRESENT, Compare object state across time (e.g., trends, changes).} \\ \\
\texttt{Ask yourself: If I want to find object \textcolor{green!50!black}{\{y\}}, does recalling past memory first help?} \\
\end{tcolorbox}

\begin{tcolorbox}[title=Prompt 4 (Continued),
colback=gray!1!white, colframe=blue!50!white, fonttitle=\bfseries, coltitle=black]
\texttt{Will it reduce time and energy needed to explore the present?} \\
\texttt{If the question involves the current state, consider both past and present.} \\ \\
\texttt{Preferred strategy order: PAST\_ONLY $>$ PAST\_THEN\_PRESENT $>$ PRESENT\_ONLY $>$ MULTI\_PAST\_AND\_PRESENT} \\
\texttt{If selecting PAST\_THEN\_PRESENT or MULTI\_PAST\_AND\_PRESENT, choose at most 5 past time instances.} \\ \\
\texttt{Based on your selected strategy, propose the ordered list of time instances to search that are most relevant.} \\ \\
\texttt{For context, here is the summary of the agent’s working memory and observations so far:} \\
\texttt{\textcolor{green!50!black}{\{working\_memory\}}}
\end{tcolorbox}

\subsection{Implementation Details Planning over Areas}
\label{appendixB2:areas}
\ph{Extended formulation of planning over areas $v$}
We adopt the object search planning formulation over a scene graph~\cite{Ginting2024Seek} to minimize the expected total path length required to find the target object $y$.
This problem is modeled as a Markov Decision Process (MDP), defined by the tuple $(S, A, T, C)$, representing the state space, action space, transition probabilities, and cost function, respectively.

The state space $S$ consists of all possible areas $v \in G_i$ where the robot can be located, along with a terminal state $s^y$ representing successful detection of object $y$.
The action space $A$ includes all areas the robot can explore.
The transition function $T$ includes both the transition probabilities between areas and the probabilities of transitioning to the goal state.
The transition probabilities to the goal state are estimated using an LLM, which is queried with the following prompt.

The cost function $C$ assigns a cost to each action, which corresponds to the path length from the robot’s current pose to the centroid of the target area, or the distance between areas.
In the case of past image retrieval, each action incurs a constant cost.

We search for a sequence of areas $[v_i, v_{i+1}, \ldots]$ that minimizes the expected cost-to-go $J$.
In our implementation, we perform a forward search up to three steps ahead to select the next area to explore, and return the first area $v_i$ in the optimal sequence.

We use the following prompt template to get the transition probabilities.
\begin{tcolorbox}
[title=Prompt 5: LLM query to get transition probabilities to find the object $y$,
colback=gray!1!white, colframe=blue!50!white, fonttitle=\bfseries,coltitle=black]
\texttt{You are an AI agent in an environment. Your task is to answer questions from the user by either exploring the environment or recalling past relevant information.} \\ \\
\texttt{To locate the object: \textcolor{green!50!black}{\{y\_object\_to\_search\}}, and to answer the question: \textcolor{green!50!black}{\{Question\}}, you must assess the probability (from 0.0 to 0.99) of finding the object in each area.} \\ \\
\texttt{Assign a probability score to each area. Higher values mean greater confidence that the object is located in that area.} \\ \\
\texttt{Only provide up to 10 areas with the highest probabilities. Do not include areas with very low likelihood.} 
\end{tcolorbox}

\begin{tcolorbox}
[title=Prompt 5 (Continued),
colback=gray!1!white, colframe=blue!50!white, fonttitle=\bfseries,coltitle=black]
\texttt{Here is the list of areas in the environment:} \\
\texttt{\textcolor{green!50!black}{\{list\_of\_area\_names\}}} \\ \\
\texttt{We are currently exploring the environment at world instance: \textcolor{green!50!black}{\{G\_i\_world\_instance\_to\_explore\}}} \\ \\
\texttt{For context, here is the agent's working memory across all world instances:} \\
\texttt{\textcolor{green!50!black}{\{working\_memory\}}}
\end{tcolorbox}

\subsection{Implementation Details Planning over Viewpoints}
We use the following prompt template to get a list of viewpoints $w$ to explore.

\begin{tcolorbox}[title=Prompt 6: Planning over viewpoints,
colback=gray!1!white, colframe=blue!50!white, fonttitle=\bfseries,coltitle=black]
\texttt{To locate the object: \textcolor{green!50!black}{\{y\_object\_to\_search\}}, and to answer the question: \textcolor{green!50!black}{\{Question\}}, what viewpoints should be searched? List at most five.} \\ \\
\texttt{Here is information about the viewpoints in the selected area v\_i. Note: the objects listed for each place are not exhaustive—they only include easily identifiable items.} \\
\texttt{\textcolor{green!50!black}{\{list\_of\_viewpoints\_in\_area\}}}
\end{tcolorbox}

The list of the viewpoints are then passed to a classical path planner and in real-world experiment, passed to a legged-robot navigation policy.

\subsection{Early Stopping of Memory Retrieval for Navigation}
\ph{Preliminaries on Value of Information}
The Value of Information (VoI) quantifies how much observing additional variables is expected to improve utility, or equivalently, reduce the expected cost-to-go $J$~\cite{howard2007information}.
Let $J^(o)$ denote the expected cost of an optimal action given observation $o$.
Then, the VoI of observing a new variable $O'$ is defined as:
\begin{align}
VOI(O' \mid o) = J^(o) - \sum_{o'} P(o' \mid o) J^*(o, o').
\end{align}
This formulation captures the expected decrease in cost resulting from observing $O'$.
If observing $O'$ does not affect the optimal action, then its VoI is zero.

We apply this principle to define early stopping conditions for searching for the object $y$ in the present environment $G_0$, beginning with memory retrieval from past world instances (e.g., $[G_1, G_2, G_0]$).

\ph{Stopping Condition 1: The prediction set contains only one area}
If the prediction set (formed using the transition probability estimate $T$, see Appendix~\ref{appendixB2:areas}) contains only one area $v_i \in G_0$ where object $y$ is likely to be found, and we assume $y$ is present in the environment and the prediction set contains the true location of $y$, then no additional observation can change the target area.
In this case, the VoI of recalling past memory is zero, as it does not influence the search strategy for $y$.

\ph{Stopping Condition 2: Further memory retrieval will not improve the next area to explore}
Consider a prediction set ${v_1, v_2, v_3}$, where $v_1$ is the closest area to the robot, and reaching $v_2$ or $v_3$ would require passing through $v_1$.
In this case, recalling additional past information will not change the robot’s next area to explore.
Thus, the VoI of retrieving more memory is zero, as it does not help prioritize the next search area for $y$.

\subsection{Mind Palace Exploration algorithm}
\begin{algorithm}[H]
\caption{Mind Palace Exploration}\label{alg:mindpalace}
\begin{algorithmic}[1]
\Require Question $Q$, episodic memories $M$, environment $E$, initial robot pose $x_0$
\State \texttt{\textcolor{green!50!black}{\# Step 1: Robotic Mind Palace generation}}
\State $\mathcal{M} \gets \texttt{mind\_palace\_generation}(M, E)$

\State \texttt{\textcolor{green!50!black}{\# Step 2: Working Memory and state initialization}}
\State $h_k \gets [\ ]$ 
\State $x_k \gets x_0$
\State $a_k \gets \texttt{null}$
\State \texttt{\textcolor{green!50!black}{\# Step 3: Loop until ready to answer}}
\State \textbf{repeat}
\State \quad \texttt{\textcolor{green!50!black}{\# Mind Palace Reasoning}}
\State \quad \textbf{if} $\texttt{is\_ready\_to\_answer}(h_k, Q)$:
\State \quad\quad $a_k \gets a^T$
\State \quad\quad \textbf{break}
\State \quad $y \gets \texttt{identify\_target\_object}(h_k, Q)$

\State \quad \texttt{\textcolor{green!50!black}{\# Mind Palace Planning}}
\State \quad $G \gets \texttt{planning\_over\_world\_instances}(y, \mathcal{M}, Q, h_k)$

\State \quad \textbf{for each} $G_i \in G$:
\State \quad\quad $v \gets \texttt{planning\_over\_areas}(y, G_i, Q, h_k)$

\State \quad\quad \textbf{for each} $v_i \in v$:
\State \quad\quad\quad $bool\_object\_found \gets \texttt{False}$
\State \quad\quad\quad $w \gets \texttt{planning\_over\_viewpoints}(y, v_i, Q, h_k)$

\State \quad\quad\quad \textbf{for each} $w_i \in w$:
\State \quad\quad\quad\quad \textbf{if} $G_i == G_0$:
\State \quad\quad\quad\quad\quad $o_k \gets \texttt{navigate}(w_i, E)$
\State \quad\quad\quad\quad\quad $\mathcal{M} \gets \texttt{update\_mind\_palace}(o_k, \mathcal{M})$
\State \quad\quad\quad\quad \textbf{else}:
\State \quad\quad\quad\quad\quad $o_k \gets \texttt{retrieve}(w_i, \mathcal{M})$

\State \quad\quad\quad\quad $bool\_object\_found \gets \texttt{is\_y\_detected}(o_k)$

\State \quad\quad\quad\quad \textbf{if} $bool\_object\_found$:
\State \quad\quad\quad\quad\quad $h_k \gets \texttt{update\_working\_memory}(o_k, h_k, G_i, v_i, w_i)$
\State \quad\quad\quad\quad\quad \textbf{break}

\State \quad\quad \textbf{if} $bool\_object\_found$:
\State \quad\quad\quad \textbf{break}

\State \quad \textbf{if} $a_k == a^T$:
\State \quad\quad \textbf{break}

\State \textbf{until} $a_k == a^T$

\State \Return $\texttt{answer\_the\_question}(Q, h_k)$
\end{algorithmic}
\end{algorithm}

%% file: Appendix/C_Long_Term_Active_EQA_Benchmarks.tex
\newpage
\section{Long-term Active EQA Benchmark}\label{appendix:benchmark-details}
The long-term active EQA benchmark consists of 150 questions annotated by a human team across five different environments.
Each environment contains 30 questions that uniformly cover all five question types. 
This section present samples of the environment and the questions.


\subsection{Habitat HM3D House 1}

\begin{figure*}[h]
    \centering
    \includegraphics[width=\linewidth]{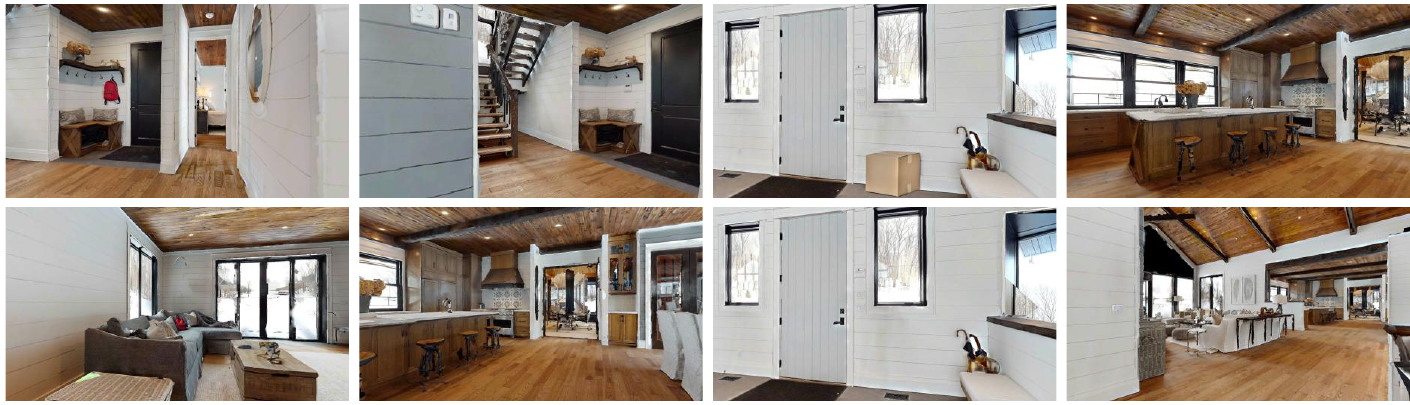}
    \caption{Sample images from the Habitat HM3D House 1.}
    \label{fig_app:habitat_1}
\end{figure*}

\begin{table}[h]
\centering
\footnotesize
\begin{tabular}{@{}p{0.52\linewidth}p{0.15\linewidth}p{0.22\linewidth}@{}}
\toprule
\textbf{Question} & \textbf{Type} & \textbf{Answer} \\
\midrule
\rowcolor{white}
What drink do we have at the kitchen counter now? & Present & Orange juice \\
\rowcolor{gray!10}
Can you check if the package that arrived on Thursday afternoon is still located at the same place? & Past-present & Yes, the package is still on the same place \\
\rowcolor{white}
What happens if I leave the drink on the kitchen island for a few days? Can I still drink it after 3 days? & Past-present-future & The orange juice will be spoiled \\
\rowcolor{gray!10}
Did I leave the orange juice on the kitchen counter overnight? & Past & No you did not \\
\rowcolor{white}
How long has the package been near the front door? & Multi-past & 2 days since Thursday afternoon \\
\bottomrule
\end{tabular}
\vspace{0.1cm}
\caption{Example questions for the Habitat HM3D House 1.}
\label{tab:temporal_questions}
\end{table}

\begin{table}[h]
\centering
\footnotesize
\begin{tabular}{@{}cl@{}}
\toprule
\textbf{\#} & \textbf{Episodic Time Instance} \\
\midrule
\rowcolor{white}
1 & Friday Afternoon \\
\rowcolor{gray!10}
2 & Thursday Afternoon \\
\rowcolor{white}
3 & Thursday Morning \\
\rowcolor{gray!10}
4 & Wednesday Afternoon \\
\rowcolor{white}
5 & Wednesday Morning \\
\rowcolor{gray!10}
6 & Tuesday Afternoon \\
\rowcolor{white}
7 & Tuesday Morning \\
\rowcolor{gray!10}
8 & Monday Afternoon \\
\rowcolor{white}
9 & Monday Morning \\
\rowcolor{gray!10}
10 & Sunday Afternoon \\
\bottomrule
\end{tabular}
\vspace{0.04cm}
\caption{List of past episodic memories from the Habitat HM3D House 1.}
\label{tab:episodic_time_instances}
\end{table}

\clearpage

\subsection{Habitat HM3D House 2}

\begin{figure*}[h]
    \centering
    \includegraphics[width=\linewidth]{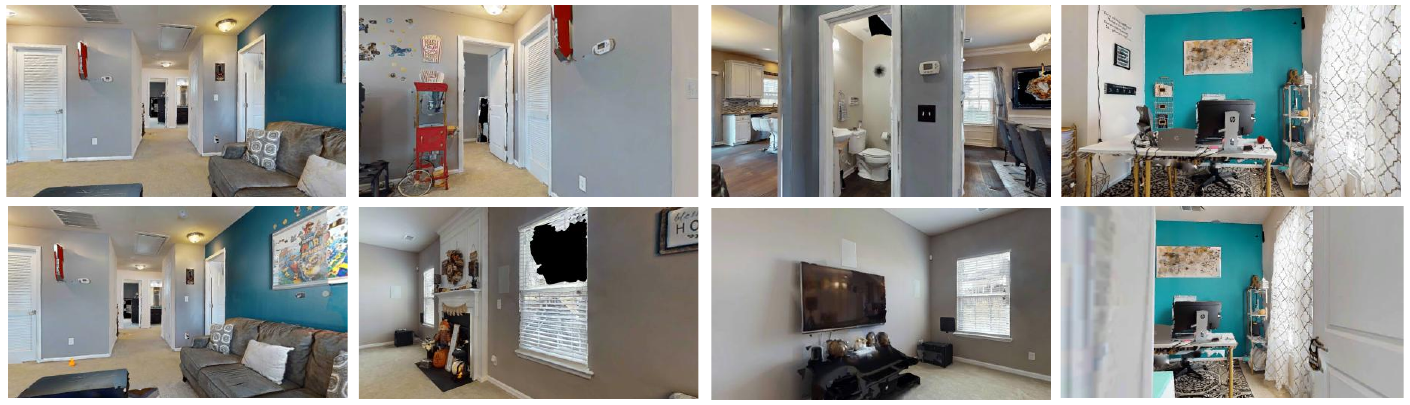}
    \caption{Sample images from the Habitat HM3D House 2.}
    \label{fig_app:habitat_2}
\end{figure*}

\begin{table}[h]
\centering
\footnotesize
\begin{tabular}{@{}p{0.52\linewidth}p{0.15\linewidth}p{0.22\linewidth}@{}}
\toprule
\textbf{Question} & \textbf{Type} & \textbf{Answer} \\
\midrule
\rowcolor{white}
What toys currently tucked on the stairways? & Present & A duck toy \\
\rowcolor{gray!10}
I'm looking for a water bottle, where is it? & Past-present & It's still on the desk in the study room \\
\rowcolor{white}
I'm going to the grocery, what kind of fruit we should buy for the house & Past-present-future & Apples, I saw apples in different places \\
\rowcolor{gray!10}
Where was the apple placed after the meal yesterday & Past & On the table in the Dining Room \\
\rowcolor{white}
Where have the kids left the duck toys for the past days? & Multi-past & On the floor in the living room and on the stair \\
\bottomrule
\end{tabular}
\vspace{0.1cm}
\caption{Example questions for the Habitat HM3D House 2.}
\label{tab:temporal_questions_house2}
\end{table}

\begin{table}[h]
\centering
\footnotesize
\begin{tabular}{@{}cl@{}}
\toprule
\textbf{\#} & \textbf{Episodic Time Instance} \\
\midrule
\rowcolor{white}
1 & Thursday Afternoon \\
\rowcolor{gray!10}
2 & Wednesday Afternoon \\
\rowcolor{white}
3 & Tuesday Afternoon \\
\rowcolor{gray!10}
4 & Monday Afternoon \\
\bottomrule
\end{tabular}
\vspace{0.04cm}
\caption{List of past episodic memories from the Habitat HM3D House 2.}
\label{tab:house2_episodic_memories}
\end{table}

\clearpage

\subsection{NVIDIA Isaac Large Warehouse}
\begin{figure*}[h]
    \centering
    \includegraphics[width=\linewidth]{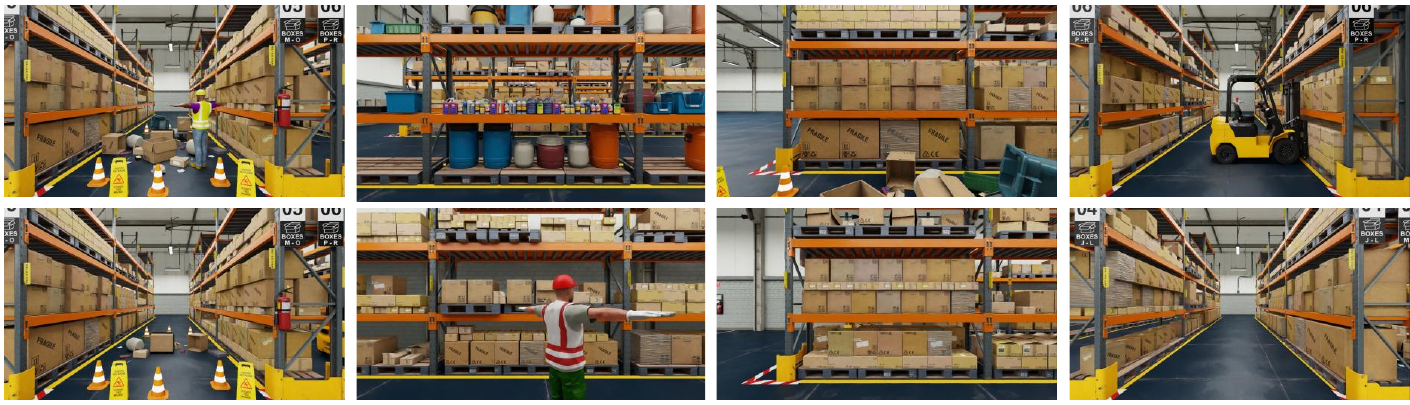}
    \caption{Sample images from the NVIDIA Isaac Large Warehouse.}
    \label{fig_app:warehouse}
\end{figure*}
\begin{table}[h]
\centering
\footnotesize
\begin{tabular}{@{}p{0.52\linewidth}p{0.15\linewidth}p{0.22\linewidth}@{}}
\toprule
\textbf{Question} & \textbf{Type} & \textbf{Answer} \\
\midrule
\rowcolor{white}
How many blue barrels do we have on the shelves in Aisle 2? & Present & 7 \\
\rowcolor{gray!10}
Can you check if the clutter has been partially cleaned up since yesterday? & Past-present & Yes, some clutter has been removed today \\
\rowcolor{white}
We want to recruit more workers in the warehouse. Which day should we assign them? & Past-present-future & There were no workers on Tuesday \\
\rowcolor{gray!10}
Was there any person close to any hazards yesterday? & Past & Yes, in Aisle 5 \\
\rowcolor{white}
Can you check if the fire extinguisher in Aisle 1 for the past days? & Multi-past & It's been placed at the same place \\
\bottomrule
\end{tabular}
\vspace{0.1cm}
\caption{Example questions for the NVIDIA Isaac Large Warehouse.}
\label{tab:temporal_questions_warehouse}
\end{table}

\begin{table}[h]
\centering
\footnotesize
\begin{tabular}{@{}cl@{}}
\toprule
\textbf{\#} & \textbf{Episodic Time Instance } \\
\midrule
\rowcolor{white}
1 & Wednesday \\
\rowcolor{gray!10}
2 & Tuesday \\
\rowcolor{white}
3 & Monday \\
\bottomrule
\end{tabular}
\vspace{0.04cm}
\caption{List of past episodic memories from the NVIDIA Isaac large warehouse.}
\label{tab:isaac_warehouse_memories}
\end{table}

\clearpage

\subsection{Real-world Office Environment}

\begin{figure*}[h]
    \centering
    \includegraphics[width=\linewidth]{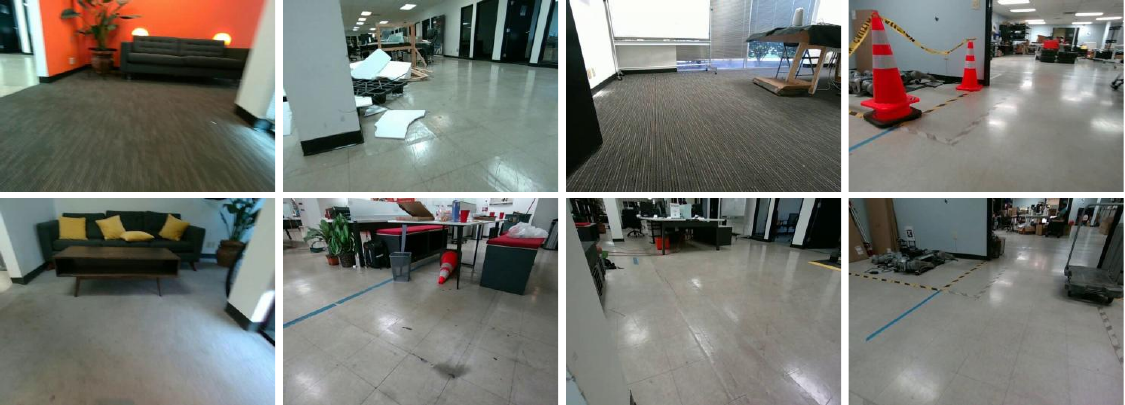}
    \caption{Sample images from the real-world office environment.}
    \label{fig_app:office}
\end{figure*}

\begin{table}[h]
\centering
\footnotesize
\begin{tabular}{@{}p{0.52\linewidth}p{0.15\linewidth}p{0.22\linewidth}@{}}
\toprule
\textbf{Question} & \textbf{Type} & \textbf{Answer} \\
\midrule
\rowcolor{white}
Is the door from the PPE room to outside closed or open? & Present & It's closed \\
\rowcolor{gray!10}
The reception area wall of the office looks different from the one I last remembered in November, what’s the difference? & Past-present & The wall is now painted orange \\
\rowcolor{white}
Can you check if the office side door through the PPE room is open? Do we usually keep it closed? & Past-present-future & We usually keep it open \\
\rowcolor{gray!10}
Do you remember the type of the orange car parked on the side of the road in January? & Past & Land Rover \\
\rowcolor{white}
What is the fire extinguisher color that we placed in the robot room in December and January? & Multi-past & Red \\
\bottomrule
\end{tabular}
\vspace{0.1cm}
\caption{Example questions for the real-world office environment.}
\label{tab:temporal_questions_office}
\end{table}

\begin{table}[h]
\centering
\footnotesize
\begin{tabular}{@{}cl@{}}
\toprule
\textbf{\#} & \textbf{Episodic Time Instance} \\
\midrule
\rowcolor{white}
1 & October 18, 2024 \\
\rowcolor{gray!10}
2 & November 15, 2024 \\
\rowcolor{white}
3 & December 13, 2024 \\
\rowcolor{gray!10}
4 & January 17, 2025 \\
\rowcolor{white}
5 & February 14, 2025 \\
\rowcolor{gray!10}
6 & March 14, 2025 \\
\rowcolor{white}
7 & March 17, 2025 (Four days ago)\\
\rowcolor{gray!10}
8 & March 18, 2025 (Three days ago)\\
\rowcolor{white}
9 & March 19, 2025 (Two days ago)\\
\rowcolor{gray!10}
10 & March 20, 2025 (Yesterday) \\
\bottomrule
\end{tabular}
\vspace{0.04cm}
\caption{List of past episodic memories from the real-world office environment.}
\label{tab:office_episodic_memories}
\end{table}

\clearpage

\subsection{Real-world High-rise Construction Site}

\begin{figure*}[h]
    \centering
    \includegraphics[width=\linewidth]{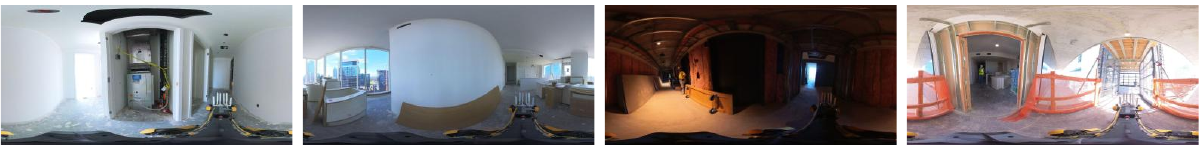}
    \caption{Sample images from the real-world high-rise construction site.}
    \label{fig_app:highrise}
\end{figure*}

\begin{table}[h]
\centering
\footnotesize
\begin{tabular}{@{}p{0.52\linewidth}p{0.15\linewidth}p{0.22\linewidth}@{}}
\toprule
\textbf{Question} & \textbf{Type} & \textbf{Answer} \\
\midrule
\rowcolor{white}
What tools are on the wooden boxes? & Present & Drilling machines \\
\rowcolor{gray!10}
Do you see the blue tape on the floor? Is it broken anywhere? & Past-present & Yes \\
\rowcolor{white}
What is the function of the blue tape used on the floor? & Past-present-future & To mark boundaries or safety zones \\
\rowcolor{gray!10}
What was the color of the fan on the ground? & Past & White \\
\rowcolor{white}
Is the yellow cart loaded or empty? & Multi-past & Empty \\
\bottomrule
\end{tabular}
\vspace{0.1cm}
\caption{Example questions for the real-world high-rise construction site.}
\label{tab:temporal_questions_construction}
\end{table}

\begin{table}[h]
\centering
\footnotesize
\begin{tabular}{@{}cl@{}}
\toprule
\textbf{\#} & \textbf{Episodic Time Instance} \\
\midrule
\rowcolor{white}
1 & August 16, 2024 \\
\bottomrule
\end{tabular}
\vspace{0.04cm}
\caption{List of past episodic memories from the real-world high-rise construction site.}
\label{tab:office_episodic_memories}
\end{table}

%% file: Appendix/D_Baseline_Details.tex
\newpage
\section{EQA Methods Details}\label{appendix:baseline-setup}

\begin{table}[h]
\centering
\footnotesize
\resizebox{\linewidth}{!}{%
\begin{tabular}{lcccccccc}
\toprule
Methods & \textbf{Poses} & \textbf{Past Img.} & \textbf{Pres. Img.} & \textbf{Past Cap.} & \textbf{Pres. Cap.} & \textbf{SG Cap.} & \textbf{Expl.} & \textbf{Full Cov.} \\
\midrule
Mind Palace (Ours) & \color{green!50!black}\checkmark & \color{green!50!black}\checkmark & \color{green!50!black}\checkmark & \color{green!50!black}\checkmark & \color{red!80!black}\ding{55} & \color{green!50!black}\checkmark & \color{green!50!black}\checkmark & \color{red!80!black}\ding{55} \\
Multi-Frame VLMs~\cite{majumdar2024openeqa} & \color{green!50!black}\checkmark & \color{green!50!black}\checkmark & \color{green!50!black}\checkmark & \color{red!80!black}\ding{55} & \color{red!80!black}\ding{55} & \color{red!80!black}\ding{55} & \color{red!80!black}\ding{55} & \color{green!50!black}\checkmark \\
Socratic LLMs~\cite{majumdar2024openeqa} & \color{green!50!black}\checkmark & \color{red!80!black}\ding{55} & \color{red!80!black}\ding{55} & \color{green!50!black}\checkmark & \color{green!50!black}\checkmark & \color{green!50!black}\checkmark & \color{red!80!black}\ding{55} & \color{green!50!black}\checkmark \\
ReMEmbR~\cite{anwar2024remembr} & \color{green!50!black}\checkmark & \color{red!80!black}\ding{55} & \color{red!80!black}\ding{55} & \color{green!50!black}\checkmark & \color{green!50!black}\checkmark & \color{red!80!black}\ding{55} & \color{red!80!black}\ding{55} & \color{green!50!black}\checkmark \\
Active EQA w/ Frames & \color{green!50!black}\checkmark & \color{green!50!black}\checkmark & \color{green!50!black}\checkmark & \color{red!80!black}\ding{55} & \color{red!80!black}\ding{55} & \color{red!80!black}\ding{55} & \color{green!50!black}\checkmark & \color{red!80!black}\ding{55} \\
Active Socratic EQA & \color{green!50!black}\checkmark & \color{red!80!black}\ding{55} & \color{green!50!black}\checkmark & \color{green!50!black}\checkmark & \color{red!80!black}\ding{55} & \color{green!50!black}\checkmark & \color{green!50!black}\checkmark & \color{red!80!black}\ding{55} \\
\bottomrule
\end{tabular}
}
\vspace{0.2cm}
\caption{\textbf{Comparison of information sources used by each method.} 
Poses = robot poses; 
Past Img. = access to past images for VLM analysis; 
Pres. Img. = access to present images for VLM analysis; 
Past Cap. = past image captions; 
Pres. Cap. = present image captions; 
SG Cap. = scene graph captions; 
Expl = supports active exploration; 
Full Cov. = requires a pre-recorded full-trajectory coverage before answering questions.}
\label{tab:comparison_infosource}
\end{table}

\subsection{Multi-Frame VLMs}
\begin{tcolorbox}[title=Prompt 7: Multi-Frame VLMs,
colback=gray!1!white, colframe=blue!50!white, fonttitle=\bfseries,coltitle=black]
\texttt{You are an AI agent in an environment. Your task is to answer questions from the user by analyzing past image observations collected by the robot.} \\ \\
\texttt{Question: \textcolor{green!50!black}{\{Question\}}} \\ \\
\texttt{Use the following information to guide your reasoning and answer the question. Do your best to answer the question based on the information you have.} \\ \\
\texttt{\textcolor{green!50!black}{\{images, robot poses\}}} \\ \\
\texttt{Think step by step.} 
\end{tcolorbox}

\subsection{Socratic LLMs}
\begin{tcolorbox}[title=Prompt 8: Socratic LLMs,
colback=gray!1!white, colframe=blue!50!white, fonttitle=\bfseries,coltitle=black]
\texttt{You are an AI agent in an environment. Your task is to answer questions from the user by analyzing past image observations collected by the robot.} \\ \\
\texttt{Question: \textcolor{green!50!black}{\{Question\}}} \\ \\
\texttt{Use the following information to guide your reasoning and answer the question. Do your best to answer the question based on the information you have.} \\ \\
\texttt{\textcolor{green!50!black}{\{image captions, scene graph captions, robot poses\}}} \\ \\
\texttt{Think step by step.} 
\end{tcolorbox}

\subsection{ReMEmbR}
We use the open-source code of ReMEmbR~\cite{anwar2024remembr} that uses GPT4o for image captioning and LLM query. 
Following their real-world setup, ReMEmbR explores all the viewpoints in the present environment to build the memory prior to the question answering. 


\subsection{Active EQA w/ Frames}
\begin{tcolorbox}[title=Prompt 9: Active EQA w/ Frames,
colback=gray!1!white, colframe=blue!50!white, fonttitle=\bfseries,coltitle=black]
\texttt{You are an AI agent in an environment. Your task is to answer questions from the user by analyzing past image observations collected by the robot.} \\ \\
\texttt{Question: \textcolor{green!50!black}{\{Question\}}} \\ \\
\texttt{Use the following information to guide your reasoning and answer the question or if you think the question requires exploration of the present state of the environment, list some viewpoints for the robot to explore in the present environment to be able to answer the question.} \\ \\
\texttt{\textcolor{green!50!black}{\{images, robot poses\}}} \\ \\
\texttt{Think step by step.} 
\end{tcolorbox}

\subsection{Active Socratic EQA}
\begin{tcolorbox}[title=Prompt 10: Active Socratic EQA,
colback=gray!1!white, colframe=blue!50!white, fonttitle=\bfseries,coltitle=black]
\texttt{You are an AI agent in an environment. Your task is to answer questions from the user by analyzing past image observations collected by the robot.} \\ \\
\texttt{Question: \textcolor{green!50!black}{\{Question\}}} \\ \\
\texttt{Use the following information to guide your reasoning and answer the question or if you think the question requires exploration of the present state of the environment, list some viewpoints for the robot to explore in the present environment to be able to answer the question.} \\ \\
\texttt{\textcolor{green!50!black}{\{image captions, scene graph captions, robot poses\}}} \\ \\
\texttt{Think step by step.} 
\end{tcolorbox}

\subsection{Implementation Details and Parameters}
All approaches and ours use the GPT-4o as the language and vision model~\cite{hurst2024gpt} and for image captioning. 
In the experiments, the maximum budget of past image retrieval is 100 images and the maximum limit of the exploration is 25 viewpoints. 

%% file: Appendix/E_Evaluation_Metrics.tex
\section{Evaluation Metrics}\label{appendix:evaluation-metrics}

We describe the three metrics used to evaluate robotic agents' performance in LA-EQA: 1) Answer correctness (Appendix~\ref{app:E1}), 2) Exploration efficiency (Appendix~\ref{app:E2}), and 3) Memory retrieval efficiency (Appendix~\ref{app:E3}).

\subsection{Answer correctness}
\label{app:E1}
The \textit{answer correctness} metric evaluates how closely the agent's answer $A$ matches the human-annotated answer $A^*$.
Since answers to EQA questions are often open-ended, there may be no single exact string that defines the correct answer, and multiple valid variations can exist.
To account for this, we adopt the \textit{LLM-Match} evaluation procedure introduced in the OpenEQA benchmark~\cite{majumdar2024openeqa}, which uses an LLM to score answer correctness.

This approach is faster and more cost-effective than manual human reviews and has demonstrated a high level of agreement with human judgments~\cite{majumdar2024openeqa}.
Given a question $Q$, a human-annotated answer $A^*$, and the agent’s answer $A$, the GPT-4o model is prompted to assign an integer score $\sigma$ from 1 to 5, where 1 represents an incorrect answer and 5 represents a fully correct response.

The answer correctness score is then normalized to a percentage:
\begin{align}
\mathcal{C} = \frac{\sigma-1}{4} \times 100\%.
\end{align}
The overall answer correctness in the benchmark is the average of $\mathcal{C}$ over all the questions. 

\begin{tcolorbox}[title=Prompt 11: Prompt used for LLM-Match scoring,
colback=gray!1!white, colframe=blue!50!white, fonttitle=\bfseries,coltitle=black]
\texttt{You are an AI agent. Your task is to evaluate the response given a question and the correct answer annotated by a human.} \\ \\
\texttt{To mark the response, output a single integer from 1 to 5 (inclusive).} \\ \\
\texttt{Question: \textcolor{green!50!black}{\{Question\}}} \\ \\
\texttt{Correct answer: \textcolor{green!50!black}{\{GT\_A\_answer\}}} \\ \\
\texttt{Response: \textcolor{green!50!black}{\{A\_answer\}}}
\end{tcolorbox}

\subsection{Exploration efficiency}
\label{app:E2}
The \textit{exploration efficiency} metric measures how efficient the agent's path to gather necessary information in the environment to answer the question correctly. 
We compare the total path length of the agent's from the start of the LA-EQA task until it answers the question against a human-annotated solution. 
The agent's total path length $p$ is measured by calculating the total path distance from the starting pose $x_0$ through the sequence of the viewpoints $w$ explored by the agent until it answers the question. 
The human annotated path length $l$ is measured by calculating the total distance to a sequence of viewpoints $w^*$ from $x_0$ that are annotated by humans that are sufficient to answer the questions.

Not all questions in the LA-EQA benchmark require present-environment exploration—for instance, questions that involve only past states. For such cases, the human-annotated sequence is set to empty, and $l = 0$.
Furthermore, the human-annotated sequence $w^*$ is not necessarily the only valid path to answer a question. For example, when annotating a solution for a question that requires checking objects on a dining table, a human may select the nearest viewpoint that provides a clear view, even though alternative, more distant or partially occluded viewpoints might also contain sufficient information.
Thus, it is possible for the agent to find a shorter valid path than the annotated reference.

Given the agent's total path length $p$ and the human-annotated path length $l$, we adapt the Success weighted by (normalized) Path Length (SPL) metric—commonly used to evaluate navigation efficiency~\cite{anderson2018evaluation,majumdar2024openeqa}—to the LA-EQA setting:
\begin{align}
\mathcal{X} =
\begin{cases}
\displaystyle \frac{\sigma - 1}{4} \times 100\%, & \text{if } l = 0 \text{ and } p = 0, \\
\displaystyle \frac{\sigma - 1}{4} \times \frac{l}{\max(l, p)} \times 100\%, & \text{otherwise.}
\end{cases}
\end{align}




\subsection{Memory Retrieval efficiency}
\label{app:E3}
The \textit{memory retrieval efficiency} metric measures the number of images retrieved by the agent to answer a given question. Images generally contain richer information than captions for answering different type of questions. 
Measuring image retrieval efficiency is particularly important in long-term memory settings, as images represent larger amount of data that must be retrieved from memory storage and processed by a VLM compared to texts. 
We count the number of images retrieved after the agent receives the question.

%% file: Appendix/F_Benchmark_Results.tex
\clearpage
\newpage
\section{Example Results from Different Methods}\label{appendix:benchmark-more-results}

\subsection{Example results from different methods}
\begin{table}[h]
\renewcommand{\arraystretch}{1.05}
\centering
\footnotesize
\begin{tabular}{|p{3.5cm}|p{3.3cm}|p{3.3cm}|p{3.3cm}|}
\hline
\multicolumn{4}{|p{13.4cm}|}{%
\textbf{Question:} Do you know where my red backpack is? I forget where I left it.
} \\
\multicolumn{4}{|p{13.4cm}|}{%
\textbf{Correct answer:} On the sofa in the living room downstairs.
} \\
\hline
\textbf{} & \textbf{Ours} & \textbf{Multi-frame VLM} & \textbf{Socratic LLMs} \\
\hline
\textbf{Retrieval and exploration} & \textcolor{green!40!black}{\textbf{Retrieve 15 images and active exploration for 6.6 m}} & Retrieve 100 images and \textcolor{red!70!black}{explore all viewpoints for 24.5 m} & \textcolor{red!70!black}{Explore all viewpoints for 24.5 m} \\
\hline
\textbf{Answer} & \textcolor{green!40!black}{Your red backpack is on the couch in the living room downstairs.} & Your red backpack is likely in the entryway. & Your red backpack was last seen in the hallway on a bench. \\
\hline
\textbf{Answer correctness} & \textcolor{green!40!black}{\textbf{100\%}} & \textcolor{orange!70!black}{25\%} & \textcolor{orange!70!black}{25\%} \\
\hline
\textbf{Exploration Efficiency} & \textcolor{green!40!black}{\textbf{54.3\%}} & 3.6\% & 3.6\% \\
\hline
\textbf{Images Retrieved} & \textcolor{green!40!black}{15 images} & \textcolor{red!70!black}{100 images} & 0 image \\
\hline
\textbf{Reasoning} & The exploration of the living room on Saturday morning shows the red backpack on the couch. This matches the past observation from Friday afternoon, confirming its current location. & The red backpack was last seen hanging in the entryway on Thursday afternoon. It was not visible in the images from Friday afternoon or Saturday morning, suggesting it might have been moved or is not in the current view. However, since it was last seen there, it is a good place to start looking. & The red backpack was observed multiple times in the hallway area, specifically on a bench, during the episodes from Thursday and Wednesday. It was not seen in the Saturday morning episode, suggesting it might still be in the same location where it was last observed. \\
\hline
\textbf{} & \textbf{ReMEmbR} & \textbf{Active EQA w/ Frames} & \textbf{Active Socratic EQA} \\
\hline
\textbf{Retrieval and exploration} & \textcolor{red!70!black}{Explore all viewpoints for 24.5 m} &Retrieve 100 images and explore viewpoints for 21.24 m &  \\
\hline
\textbf{Answer} & Your red backpack is hanging on a hook in the hallway with wooden stairs and a black door. & I couldn't find your red backpack in the explored areas. It might have been moved to a different location. & Your red backpack was last seen in the hallway areas with benches and hooks. \\
\hline
\textbf{Reasoning} & Your red backpack is hanging on a hook in the hallway with wooden stairs and a black door. & The red backpack was last seen in the hallway near the stairs on Thursday afternoon. The images from Saturday morning show the hallway and the stairs, but the backpack is not visible in any of these images. This suggests that the backpack has been moved from its last known location. Since the images explored do not show the backpack, it is likely in a different area not covered by the current exploration. & The red backpack was seen in multiple places during the week, but most recently on Thursday afternoon, it was observed in several locations such as Place Node 18, Place Node 21, Place Node 24, Place Node 27, Place Node 30, Place Node 33, Place Node 36, and Place Node 39. Since the user mentioned they forgot where they left it last night, it is likely still in one of these locations. \\
\hline
\textbf{Answer correctness} & \textcolor{red!70!black}{0\%} & \textcolor{orange!70!black}{25\%} & \textcolor{orange!70!black}{25\%} \\
\hline
\textbf{Exploration efficiency} & 0\% & 4.2\% & 25\% \\
\hline
\textbf{Images Retrieved} & 0 image & \textcolor{red!70!black}{100 images} & 0 image \\
\hline
\end{tabular}
\vspace{0.2cm}
\caption{Example results from different methods}
\end{table}

\clearpage
\newpage
\subsection{Example comparison on early stopping}

\begin{table}[h]
\renewcommand{\arraystretch}{1.2}
\centering
\footnotesize
\begin{tabular}{|p{3.5cm}|p{4.5cm}|p{4.5cm}|}
\hline
\multicolumn{3}{|p{12.5cm}|}{%
\textbf{Question:} Do you know where my red backpack is? I forget where I left it.
} \\
\multicolumn{3}{|p{12.5cm}|}{%
\textbf{Correct answer:} On the sofa in the living room downstairs.
} \\
\hline
\textbf{} & \textbf{Ours} & \textbf{Ours w/ early stopping} \\
\hline
\textbf{Retrieval and exploration} & Retrieve 15 images (Friday afternoon, thursday afternoon, wednesday afternoon), and
active exploration for 6.6 m & Retrieve 5 images (Friday afternoon),
\textcolor{green!40!black}{\textbf{early stopping of image retrieval (not searching thursday afternoon, wednesday afternoon)}}, and 
active exploration for 5.7 m \\
\hline
\textbf{Answer} & Your red backpack is on the couch in the living room downstairs. & The red backpack is on the couch in the living room downstairs. \\
\hline
\textbf{Reasoning} & The exploration of the living room on Saturday morning shows the red backpack on the couch. This matches the past observation from Friday afternoon, confirming its current location. & The exploration of both past and present environments consistently shows the red backpack on the couch in the living room downstairs. The images from both Friday afternoon and Saturday morning confirm this location. \\
\hline
\textbf{Answer Correctness} & 100\% & 100\% \\
\hline
\textbf{Exploration Efficiency} & 54.3\% & \textcolor{green!40!black}{\textbf{61.8\%}} \\
\hline
\textbf{Images Retrieved} & 15 images  & \textcolor{green!40!black}{\textbf{5 images}} \\
\hline
\end{tabular}
\vspace{0.2cm}
\caption{Comparison of results with and without early stopping of memory retrieval}
\end{table}

%% file: Appendix/G_Hardware_Experiments.tex
\clearpage
\newpage
\section{Hardware Experiment Details}\label{appendix:hardware-results}

\begin{figure*}[h]
    \centering
    \includegraphics[width=\linewidth]{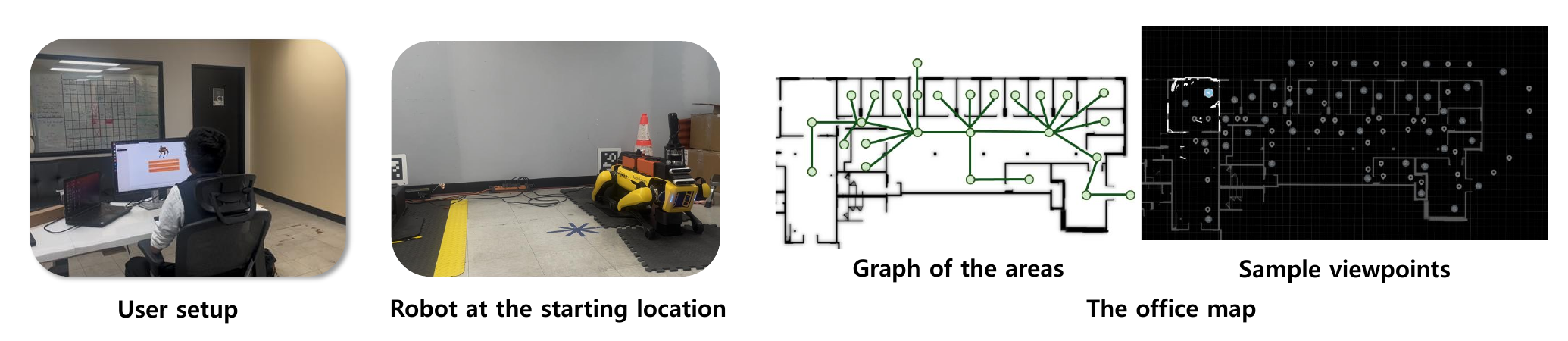}
    \caption{Hardware experiment setup in the office environment.}
    \label{fig_app:hw_setup}
\end{figure*}

\subsection{List of past episodic memories}
\begin{table}[h]
\centering
\footnotesize
\begin{tabular}{@{}clc@{}}
\toprule
\textbf{\#} & \textbf{Date} & \textbf{Distance (meters)} \\
\midrule
\rowcolor{white}
1 & October 18, 2024 & 224 \\
\rowcolor{gray!10}
2 & November 15, 2024 & 258 \\
\rowcolor{white}
3 & December 13, 2024 & 191 \\
\rowcolor{gray!10}
4 & January 17, 2025 & 503 \\
\rowcolor{white}
5 & February 14, 2025 & 492 \\
\rowcolor{gray!10}
6 & March 14, 2025 & 480 \\
\rowcolor{white}
7 & March 17, 2025 & 128 \\
\rowcolor{gray!10}
8 & March 18, 2025 & 134 \\
\rowcolor{white}
9 & March 19, 2025 & 120 \\
\rowcolor{gray!10}
10 & March 20, 2025 & 115 \\
\bottomrule
\end{tabular}
\vspace{0.04cm}
\caption{List of past episodic memories collected from robot trajectories in the office.}
\label{tab:episodic_memories}
\end{table}

\subsection{List of questions}
\begin{table}[h]
\centering
\footnotesize
\resizebox{\linewidth}{!}{%
\begin{tabular}{@{}cl@{}}
\toprule
\textbf{\#} & \textbf{Questions used for the hardware experiments} \\
\midrule
\rowcolor{white}
1 & Please inspect the fire extinguisher in the robot room and let me know if it has been in its usual location over the past few months. \\
\rowcolor{gray!10}
2 & Do you know where my jacket is? I think I left it in the meeting room yesterday, but I didn’t see it there this morning. \\
\rowcolor{white}
3 & Can you find a desk for the guest to use today? Look for one that hasn’t been used in the past few days. \\
\rowcolor{gray!10}
4 & Can you check if the office side door through the PPE room is open? Do we usually keep it closed? \\
\rowcolor{white}
5 & I’m looking for a package that was delivered. It’s a large orange case. Do you know where it is? \\
\rowcolor{gray!10}
6 & We are changing the ceiling light bulbs. Is there any equipment available that could help us? \\
\rowcolor{white}
7 & The reception area seems different from when I last visited in January. Has anything changed? \\
\bottomrule
\end{tabular}
}
\vspace{0.04cm}
\caption{List of questions used in the hardware experiments. All questions require both memory retrieval and active exploration.}
\label{tab:hardware_questions}
\end{table}

%% file: Appendix/H_Failure_Cases.tex
\section{Examples of Failure Cases}\label{appendix:failure-modes}
\begin{figure*}[h]
    \centering
    \includegraphics[width=\linewidth]{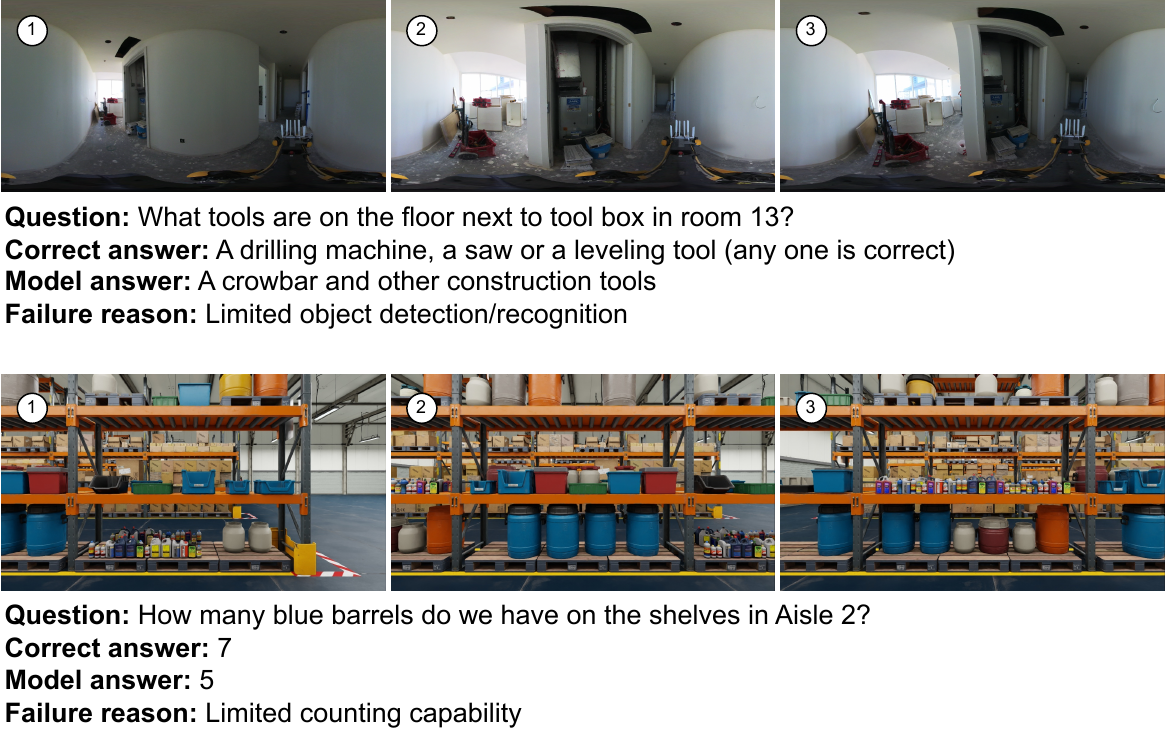}
    \caption{Examples of the failure cases.}
    \label{fig_app:failure_examples}
\end{figure*}
\autoref{fig_app:failure_examples} shows two failure cases due to limitations of the current VLMs' semantic and spatial understanding capabilities. The first example shows the failure to recognize objects in the images. Despite retrieving the correct images to answer the question, the VLM fails to return the correct answer in our experiment. 
The second example involves counting objects through multiple images. The retrieved images contain overlapping views that cover all the blue barrels in Aisle 2. However, we often fail to give the correct number of objects. 
These failure cases highlight the need for improved semantic and spatial understanding of approaches that use vision and language models to improve the performance on the LA-EQA benchmark.


%% file: Appendix/I_Full_Results.tex
\section{Example of Full Results}\label{appendix:full-examples}
\begin{tcolorbox}[title=Example 2: Full example of Mind Palace Exploration for LA-EQA (part 1/2),
colback=gray!1!white, colframe=green!50!white, fonttitle=\bfseries,coltitle=black]
\footnotesize
\texttt{Question: When was the package at the front door upstairs delivered?} \\
\texttt{Ground truth answer: Before Thursday afternoon} \\ \\

\textcolor{blue}{\texttt{\# Reasoning over LA-EQA Iteration 1}} \\
\texttt{Object to search: package at the front door upstairs} \\
\texttt{Reasoning: To determine when the package was delivered, I need to look for a package at the front door upstairs, as it is the specific object mentioned in the question.} \\ \\

\textcolor{blue}{\texttt{\# Planning over episodic world instances}} \\
\texttt{Search strategy: PAST ONLY} \\
\texttt{Reasoning on search strategy: The user question is about when the package was delivered, which is a past event. Therefore, the most efficient strategy is to recall past memory to find when the package was first observed at the front door upstairs. Exploring the present environment is not necessary as it will not provide information about the delivery time.} \\
\texttt{EpisodicExploration: Time instance to retrieve: ['friday afternoon', 'thursday afternoon', 'wednesday afternoon']} \\
\texttt{EpisodicExploration: Reasoning: These times are relevant because they are the most recent past instances before the current time. Checking these instances will help identify when the package was first observed at the front door upstairs, which is crucial to determine the delivery time.} \\ \\

\textcolor{blue}{\texttt{\# Searching for the target object in 'friday afternoon'}} \\
\textcolor{blue}{\texttt{\# Planning over areas}} \\
\texttt{Area list: ['main entrance', 'Living room upstairs', 'stairs', 'kitchen and dining', 'Living room downstairs']} \\
\texttt{Area probability: [0.85, 0.7, 0.5, 0.3, 0.2]} \\
\texttt{Reasoning: The task is to locate a package at the front door upstairs and determine when it was delivered. Given the room names and their descriptions, the main entrance is the most likely location for a package delivery, as it is typically where packages are left. The living room upstairs is also a potential location since it is upstairs and could be near the front door. The stairs could be a transitional area where the package might be temporarily placed. The kitchen and dining area, while not directly related to package delivery, could be a place where someone might move the package after receiving it. The living room downstairs is less likely since it is not upstairs, but it could still be a place where the package was moved. Based on these considerations, I assign the highest probability to the main entrance, followed by the living room upstairs, stairs, kitchen and dining, and living room downstairs.} \\ \\

\textcolor{blue}{\texttt{\# Planning over viewpoints}} \\
\texttt{Viewpoints to explore: [198, 204, 207, 201, 213]} \\
\texttt{Reasoning: To locate the package at the front door upstairs, we should search places that mention a door and a cardboard box, as these are likely locations for package delivery. Place Node 198, 204, and 207 all mention a cardboard box and a door, making them relevant to the search. Additionally, Place Node 201 and 213 mention a door and are part of the main entrance, which is a common area for package delivery.} \\
\texttt{Image retrieval…} \\
\texttt{Image Analysis Answer: Object found!} \\
\texttt{Reasoning: The images show a package at the front door upstairs, indicating its presence. However, the images do not provide information about when it was delivered.} 
\end{tcolorbox}

\begin{tcolorbox}[title=Full example of Mind Palace Exploration for LA-EQA (part 2/2),
colback=gray!1!white, colframe=green!50!white, fonttitle=\bfseries,coltitle=black]
\footnotesize
\textcolor{blue}{\texttt{\# Searching for the target object in 'thursday afternoon'}} \\
\textcolor{blue}{\texttt{\# Planning over areas (main entrance)}} \\
\textcolor{blue}{\texttt{\# Episode to explore: thursday afternoon}} \\
\textcolor{blue}{\texttt{\# Planning over viewpoints}} \\
\texttt{Image retrieval…} \\
\texttt{Image Analysis Answer: Object found!} \\
\texttt{Reasoning: The images show a package placed near the front door, indicating that it has been delivered.} \\ \\

\textcolor{blue}{\texttt{\# Searching for the target object in 'wednesday afternoon'}} \\
\textcolor{blue}{\texttt{\# Planning over areas (main entrance)}} \\
\textcolor{blue}{\texttt{\# Planning over viewpoints}} \\
\texttt{Image retrieval…} \\
\texttt{Image Analysis Answer: Object NOT found!} \\

\textcolor{blue}{\texttt{\# Planning over areas (Living room upstairs)}} \\
\textcolor{blue}{\texttt{\# Planning over viewpoints}} \\
\texttt{Image retrieval…} \\
\texttt{Image Analysis Answer: Object NOT found!} \\

\textcolor{blue}{\texttt{\# Planning over areas (stairs)}} \\
\textcolor{blue}{\texttt{\# Planning over viewpoints}} \\
\texttt{Image retrieval…} \\
\texttt{Image Analysis Answer: Object NOT found!} \\

\textcolor{blue}{\texttt{\# Planning over areas (kitchen)}} \\
\textcolor{blue}{\texttt{\# Planning over viewpoints}} \\
\texttt{Image retrieval…} \\
\texttt{Image Analysis Answer: Object NOT found!} \\
\texttt{Reasoning: The package was not found in Wednesday afternoon} \\ \\

\textcolor{blue}{\texttt{\# Reasoning over LA-EQA Iteration 2}} \\
\texttt{Ready to answer: Yes} \\
\texttt{Reasoning: The package was already present on Thursday afternoon, as seen in the images. Therefore, it was delivered before Thursday afternoon.} \\ \\

\texttt{Answer: }\textcolor{green!50!black}{\texttt{Before Thursday afternoon}} \\
\texttt{Reasoning: }\textcolor{green!50!black}{\texttt{The package was observed at the front door upstairs on Thursday afternoon. It was not present on Wednesday afternoon, indicating it was delivered sometime between Wednesday afternoon and Thursday afternoon.}}

\end{tcolorbox}